\documentclass[10pt,twocolumn,letterpaper]{article}

\usepackage[pagenumbers]{cvpr} 

\usepackage{amsmath,amssymb,amsthm}
\newtheorem{lemma}{Lemma}

\newtheorem{theorem}{Theorem}

\usepackage{xcolor} 
\definecolor{myblue}{RGB}{0,164,239}
\definecolor{mygreen}{RGB}{127,186,0}
\definecolor{myred}{RGB}{242,80,34}
\newcommand{\chuheng}[1]{{\color{cyan}#1}}

\makeatletter
\renewcommand\@fnsymbol[1]{}
\makeatother
\definecolor{cvprblue}{rgb}{0.21,0.49,0.74}
\usepackage[pagebackref,breaklinks,colorlinks,allcolors=cvprblue]{hyperref}

\def\paperID{8584} 
\def\confName{CVPR}
\def\confYear{2026}

\title{\textcolor{myblue}{\textbf{Discover}}, \textcolor{mygreen}{\textbf{Learn}}, and \textcolor{myred}{\textbf{Reinforce}}: Scaling Vision-Language-Action Pretraining with Diverse RL-Generated Trajectories}

\author{
Rushuai Yang$^{1,5 \thanks{This work was done when Rushuai Yang was an intern at Microsoft Research Asia.}}$
\quad
Zhiyuan Feng$^{2}$
\quad
Tianxiang Zhang$^{3}$
\quad
Kaixin Wang$^{5}$
\quad \\
Chuheng Zhang$^{5,\dagger}$
\quad 
Li Zhao$^{5}$
\quad
Xiu Su$^{4}$
\quad 
Yi Chen$^{1,\dagger}$
\quad
Jiang Bian$^{5}$\and 
$^{1}$The Hong Kong University of Science and Technology \quad
$^{2}$Tsinghua University \\
$^{3}$Wuhan University \quad
$^{4}$Central South University \quad
$^{5}$Microsoft Research
\thanks{$\dagger$ Correspondence to: \texttt{chuhengzhang@microsoft.com}, \texttt{yichen@ust.hk}}
}
\begin{document}
\maketitle
\begin{abstract}
Scaling vision-language-action (VLA) model pre-training requires large volumes of diverse, high-quality manipulation trajectories.
Most current data is obtained via human teleoperation, which is expensive and difficult to scale. 
Reinforcement learning (RL) methods learn useful skills through autonomous exploration, making them a viable approach for generating data.
However, standard RL training collapses to a narrow execution pattern, limiting its utility for large-scale pre-training.
We propose \textbf{Discover, Learn and Reinforce (DLR)}, an information-theoretic pattern discovery framework that generates multiple distinct, high-success behavioral patterns for VLA pretraining. 
Empirically, DLR generates a markedly more diverse trajectory corpus on LIBERO.
Specifically, it learns multiple distinct, high‑success strategies for the same task where standard RL discovers only one, and hence it covers substantially broader regions of the state–action space. 
When adapted to unseen downstream task suites, VLA models pretrained on our diverse RL data surpass counterparts trained on equal‑sized standard RL datasets. 
Moreover, DLR exhibits positive data‑scaling behavior that single‑pattern RL lacks. 
These results position multi‑pattern RL as a practical, scalable data engine for embodied foundation models.
\end{abstract}
        
\section{Introduction}
\label{sec:intro}
\begin{figure*}[t]
    \centering
    \includegraphics[width=\textwidth]{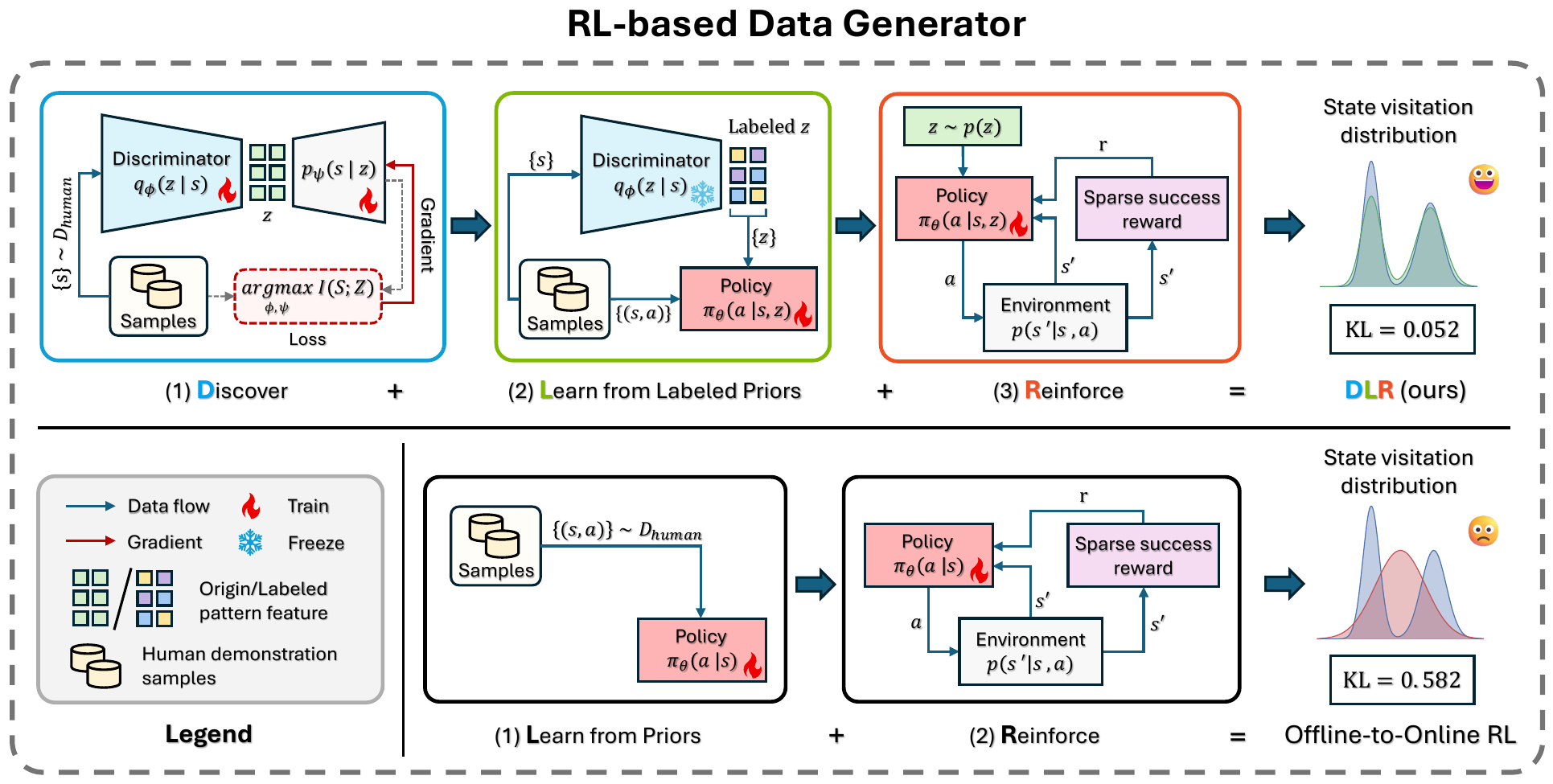}
    \caption{\textbf{Comparison between our DLR framework and a standard offline-to-online RL baseline.}
    The top row illustrates our three-stage DLR process: (1) We discover latent patterns from human data using a VAE-based approach. (2) We learn a pattern-conditioned policy via behavior cloning on the now-labeled data. (3) We refine each pattern-conditioned policy online with a sparse success reward. This process results in a diverse, multi-modal state visitation distribution.
    The bottom row shows a standard 
    offline-to-online
    RL baseline: (1) A policy is initialized via behavior cloning on the entire unlabeled human dataset. (2) The policy is refined online with a sparse success reward. This standard approach leads to mode collapse, resulting in a uni-modal state visitation distribution.}
    \label{fig:offline_to_online_rl}
\end{figure*}

Vision-language-action (VLA) models have established a dominant paradigm of large-scale pretraining followed by downstream task fine-tuning \cite{yang2025magma, zhao2025cot, lin2024vila, szot2025multimodal, li2024manipllm}. 
The goal of pre-training is to expose the model to diverse behaviors of robots for it to acquire broad manipulation capabilities. The subsequent fine-tuning phase then adapts this foundation to skillfully execute specific tasks~\cite{kim2025openvla, pi0, zitkovich2023rt, ICLR2025_49f80e4d, chen2025villa}. The efficacy of this paradigm is critically dependent on the scale and diversity of the pre-training data \cite{pi05, wen2025data, chen2025vidbot}. Currently, this data is primarily sourced from human teleoperation, a process that is not only labor-intensive and costly but also inherently limited in the behavioral diversity.
Driven by the sole goal of task success, human demonstrators naturally
rely on a few efficient strategies, rather than deliberately demonstrating alternative viable solutions~\cite{Zipf}. This limitation poses a fundamental challenge to generating the rich, multi-pattern data required for effective VLA pre-training.

Reinforcement learning (RL) has emerged as a powerful alternative for enabling robots to acquire complex patterns through environmental interaction \cite{Rajeswaran-RSS-18, janny2025reasoning, hoeller2024anymal}.
Its fundamental strength lies in the trial-and-error process: by optimizing the reward signal, an agent can autonomously discover efficient strategies that often surpass what can be learned by simply imitating human demonstrations~\cite{berges2023galactic, radosavovic2024real}. Previous work has demonstrated the potential of RL to refine VLA policies on specific tasks—achieving smoother, more efficient behaviors than human demonstrators or even discovering novel successful strategies \cite{luo2025precise, wurman2022outracing, johannink2019residual, zhang2024imagine}. 
However, prior work focuses primarily on using RL for VLA fine-tuning, while its potential for enabling VLA pre-training remains largely understudied. In this paper, we explore how to collect diverse trajectories for VLA pre-training, which is important for VLA to succeed even in a single task in the same environment~\cite{shi2025diversity,pi05}.
By design, the objective of policy-based RL is to find the optimal policy, which often leads to convergence on a fixed
execution pattern~\cite{NEURIPS2024_43d1d3bd,yang2023policy}.
While highly effective for mastering a specific skill, the resulting trajectories may not possess the diversity required to instill the rich knowledge needed for downstream generalization. Therefore, designing an RL framework capable of explicitly generating a diverse dataset for VLA pre-training emerges as a critical research challenge.

In this paper, we propose to reframe the goal of RL training to discovering a repertoire of high-success behavioral patterns for each task. Concretely, we introduce the three-stage \textcolor{myblue}{\textbf{Discover}}, \textcolor{mygreen}{\textbf{Learn}}, and \textcolor{myred}{\textbf{Reinforce}} \textbf{(\textcolor{myblue}{D}\textcolor{mygreen}{L}\textcolor{myred}{R})} framework:
We 1) \textcolor{myblue}{\textbf{discover}} distinct behavioral patterns from human demonstrations using the information-theoretic principle;
2) \textcolor{mygreen}{\textbf{learn}} a pattern-conditioned policy to imitate these discovered patterns; and
3) \textcolor{myred}{\textbf{reinforce}} the pattern-conditioned policy using the task reward towards the refined solutions corresponding to different patterns. This process yields a multi-pattern policy where each behavior serves as a high-quality data generator, enabling diverse sampling for VLA pretraining.

We conduct experiments to evaluate the out-of-distribution generalization capability using the LIBERO benchmark~\cite{liu2023libero}.
Specifically, we pre-train the VLA model on tasks from LIBERO-90 with the data collected using RL, and fine-tune the model on tasks from LIBERO-spatial/object/goal/long respectively.
Our experiment results indicate that 1) the VLA model pre-trained on the RL data with multiple patterns generated by DLR outperforms its counterpart pre-trained on an equal-sized dataset collected using canonical RL when fine-tuned on downstream tasks, and 2) the performance of VLA scales with the data volume when DLR is used to collect the data. These findings suggest the possibility to shift from human-centric to algorithmically generated data pipelines, reducing cost while enabling principled scaling. In summary, our paper makes the following contributions:
\begin{itemize}
    \item We propose a principled three-stage framework, DLR, to generate high-quality and diverse robotic trajectories for VLA pre-training using reinforcement learning.
    \item We provide a theoretical analysis to demonstrate the ability of DLR to preserve the diversity of discovered patterns and prevent collapsing to a single solution.
    \item We show that DLR can not only generate diverse successful trajectories but also result in pre-trained VLAs that perform better when fine-tuned on downstream tasks.
\end{itemize}

\section{Related Work}
\label{sec:related_work}

\subsection{Data Generation for VLA}
Diverse and high-quality trajectories are crucial in producing generalist VLA models. 
There are several approaches to collect such trajectories. 
The human-centric approach is effective and popular, including teleoperation on real robots~\cite{oxe,AgiBotWorld,behavior1k} and simulated environments~\cite{robotwin, isaacgym}. However, they are labor‑intensive and hard to scale. Another approach learns world models and generates trajectories from the learned dynamics~\cite{racaniere2017imagination, wang2024driving, bar2025navigation}.
While this approach is attractive due to its ability to generate diverse trajectories, current world models still struggle with precise, long‑horizon robot motions and suffer from accumulated error over time~\cite{finn2016unsupervised, duan2024learning, zhu2025irasim}.
A promising alternative is generating data using RL,
which treats the trained RL policy as a data generator and can exceed human demonstrations in quality~\cite{rldg, Rajeswaran-RSS-18, wang2019reinforced, choi2019robust}.
Yet standard RL is optimized purely for task success and typically converges to a single solution, even when multiple exist. For a narrow task, one solution may suffice, but it yields a poor dataset for general VLA training. In contrast, we follow the RL‑generator paradigm and explicitly induce distinct high‑success strategies, broadening the state–action coverage and providing richer priors for VLA generalization.

\subsection{Reinforcement Learning for VLA}

Previously, RL has been leveraged to improve VLA but mainly in fine-tuning or post-training. For example, online RL~\cite{hu2025flare, li2025simplevla, ChenY1-RSS-25, zhai2024fine, lu2025vla, liu2025can, chen2025tgrpo, zang2025rlinf, guo2025improving} is used for fine-tuning VLA via online interaction, but it is usually costly and slow in practice since the VLA models are typically large.
Offline RL~\cite{Brohan-RSS-23, octo_2023, yue2024deer, liu2024robomamba, zhang2025balancing, huang2025co, zhao2025more, zhang2025reinbot} avoids online exploration by training from fixed data, but it is sensitive to the quality of datasets and reward specification. Our DLR framework takes a different path: instead of optimizing large VLA models with RL, we train small policies that are only used to generate diverse and high-quality trajectories. We then use this data to pretrain the large VLA models. 
This strategy shifts the expensive RL optimization away from the large VLA model and onto a lightweight, specialized policy used only for data generation.

\section{Preliminaries: MDP and RL Notation}
\label{sec:preliminaries}
The agent-environment interaction is modeled as a Markov Decision Process (MDP) \cite{sutton1998reinforcement},
defined by the tuple \(\mathcal{M}=(\mathcal{S},\mathcal{A},\mathcal{P},\mathcal{R},\rho_0, \gamma)\) where \(\mathcal{S}\) is the state space, \(\mathcal{A}\) is the action space, \(\mathcal{P}(s'~|~s,a)\) is the state transition function, \(\mathcal{R}(s,a)\) is the reward function, \(\rho_0(s)\) is the initial state distribution, and \(\gamma \in [0,1)\) is the discount factor. 
A policy \(\pi_\theta(\cdot|s)\) with parameters \(\theta\) maps the state $s$ to a distribution over actions. The agent interacts with the environment by taking actions sampled from its policy,
generating a trajectory \(\tau=(s_0,a_0,s_1,a_1,\dots,s_T)\) where $s_0$ is sampled from the initial state distribution $\rho_0$. The discounted state visitation distribution for a policy $\pi$ is defined as
\begin{equation}
    d^\pi(s) = (1-\gamma) \sum_{t=0}^{\infty} \gamma^t \Pr(s_t=s | s_0 \sim \rho_0, \pi)
\end{equation}
where $\Pr(s_t=s | s_0 \sim \rho_0, \pi)$ is the probability of being in state $s$ at timestep $t$ after starting from the initial distribution $\rho_0$ and following the policy $\pi$.
To obtain the optimal policy that generates high-quality data, the standard RL objective is to learn a policy \(\pi_\theta\) that maximizes the expected return, formulated as:
\begin{equation}
    J(\theta) = \mathbb{E}_{\tau \sim \pi_\theta}[R(\tau)]
\end{equation}
where the discounted return for a trajectory is defined as \(R(\tau) = \sum_{t=0}^{T-1} \gamma^t r_t\), with \(\gamma \in [0, 1)\) being the discount factor. In practice, we are provided with
a sparse reward function where the agent receives the reward $+1$ only on the final step of a successful trajectory, and the reward $0$ otherwise. In this case, with \(\mathbb{I}_{\text{succ}}(\tau)\) denoting a binary indicator for a successful trajectory, we can rewrite the discounted return as $R(\tau) = \gamma^{T-1} \cdot \mathbb{I}_{\text{succ}}(\tau)$. Note that the discount factor encourages shorter (i.e., more efficient) successful trajectories.

To generate a diverse repertoire of behaviors, we introduce a latent \(z \in \mathcal{Z}\) to represent distinct behavioral patterns. 
This results in a pattern-conditioned policy, \(\pi_\theta(a~|~s,z)\) with an additional condition on the latent $z$. During inference, we sample $z$ from a prior distribution $p(z)$, typically a uniform categorical distribution where each category corresponds to a specific pattern.
By conditioning on different latent \(z\), the policy can produce diverse successful trajectories.

For the pattern-conditioned policy, it is optimized to generate a diverse repertoire of successful trajectories. Therefore, we maximize the expected return averaged over both the distribution of patterns and the trajectories sampled from the corresponding conditioned policy, i.e., 
\begin{equation}
    J(\theta) = \mathbb{E}_{z \sim p(z), \tau \sim \pi_\theta(\cdot~|~z)} [R(\tau)]\chuheng{.}
\end{equation}

\section{A Principled Framework for Diverse Trajectory Generation}
\label{sec:method_framework}
\begin{figure*}[t]
    \centering
    \includegraphics[width=0.9\textwidth]{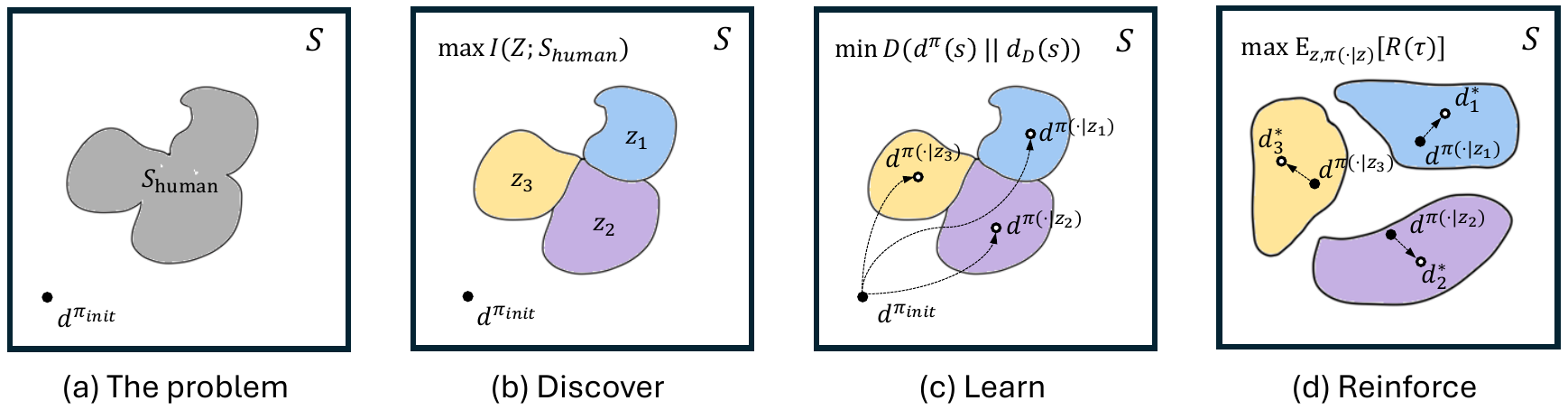}
    \caption{\textbf{The Learning Process of DLR.}
    Each panel represents the same state space \(\mathcal{S}\). The amorphous regions depict areas of visited states under different policies, and the black dots represent the corresponding state visitation distributions induced by the policy.
    \textbf{(a)} Given suboptimal human demonstrations (\(S_{\text{human}}\), gray) that cover multiple distinct successful strategies, our goal is to learn several optimal policies with high task success rates and high pattern diversity from an initial, randomly initialized policy distribution $d^{\pi_\text{init}}$.
    \textbf{(b)} In Stage 1, we discover underlying behavioral patterns from \(S_{\text{human}}\) by clustering the states into distinct modes, each identified by a latent code (\(z_1\), \(z_2\), \(z_3\)) and visualized with a different color (blue, purple, yellow).
    \textbf{(c)} In Stage 2, we use behavior cloning to train a conditional policy \(\pi(\cdot|z)\) that imitates each discovered mode. The dashed arrows indicate the cloning process, mapping from a general initial policy to more specialized ones.
    \textbf{(d)} In Stage 3, each pattern-conditioned policy is fine-tuned with sparse task rewards, letting each one converge to its own optimal version, \(d_k^*\).}
    \label{fig:three_stage_framework}
\end{figure*}
In this section, we develop a principled framework for learning different high-success behavioral patterns distinguishable by a latent variable. Our starting point is the line of work on diversity-driven reinforcement learning~\cite{diayn, dads, apt, cic}, which typically optimizes for behavior distinguishability, often using mutual information, without an explicit focus on task success.  A natural next step, which we analyze first, is to combine these diversity objectives with a standard task reward to jointly optimize for both goals. 
However, we identify a fundamental conflict in this naive combination. Such diversity objectives come with a side effect that they penalize the policy for exploring successful states, hindering its ability to discover successful solutions that pass through these states. To resolve this, our core contribution is to apply the diversity objective only to states within successful trajectories. Based on this principle, we derive a practical three-stage algorithm that can learn a pattern-conditioned policy with theoretical support for its diversity.

\subsection{Combining Task Reward with a Diversity Bonus}
\label{sec:general_infomax}
Our analysis begins with the methods designed purely for encouraging behavioral diversity. A prominent approach in prior work \cite{diayn, dads, apt, cic} is to maximize the mutual information (MI) \(I(Z;S)\) between the latent \(z\) and the states \(s\) the policy visits.  This objective encourages the policy to generate trajectories where the states $s$ are highly indicative of the underlying pattern \(z\).

To adapt this for generating successful and diverse data, a straightforward approach is to combine mutual information with the standard task performance objective:
\begin{equation}
\max_{\theta} \quad \mathbb{E}_{z\sim p(z),\ \tau\sim\pi_\theta(\cdot~|~z)}\big[ R(\tau) \big] + \beta\, I_\theta(Z;S)
\label{eq:info-rl-main}
\end{equation}
where $I_\theta(Z;S)$ is the mutual information (MI) between the latent $z \sim p$ and the state $s \sim d^{\pi_\theta (\cdot | \cdot, z)}$, and the hyperparameter $\beta \ge 0$ balances the two objectives. Intuitively, by maximizing the mutual information, different latent lead to different trajectories, as this increases the predictability of the state $s$ given the latent $z$.

However, the mutual information $I_\theta(Z;S)$ is computationally intractable. To optimize this objective,
we maximize its variational lower bound,
\begin{equation}
\label{eq:barber-agakov}
I_\theta(Z;S) \ge \mathbb{E}_{z \sim p(z), s \sim d^{\pi_\theta (\cdot | \cdot, z)}
}[\log q_\phi(z~|~s) - \log p(z)]
\end{equation}
where $q_\phi(z|s)$ parameterized by $\phi$ is the approximate posterior, and it is learned by maximum likelihood.
See Appendix~\ref{app:mi_derivation} for full derivation. This objective can be reformulated as the standard RL objective with an intrinsic reward that encourages diversity,
\begin{equation}
    \label{eq:intrinsic-reward}
    r_{\mathrm{div}}(s,z) = \log q_\phi(z~|~s) - \log p(z).
\end{equation}Then, we can optimize in an alternative procedure, where $q_\phi$ is first trained on trajectories of the current policy, and then $\pi_\theta$ is updated to optimize
\begin{equation}
    \label{eq:full-objective}
    \max_{\theta} \quad \mathbb{E}_{z, \tau, s}\left[ R(\tau) + \beta \cdot (\log q_\phi(z~|~s) - \log p(z)) \right].
\end{equation}
\subsection{The Conflict Between Task and Diversity Objectives}
\label{sec:pathology}
However, directly summing the task reward and the diversity reward in Eq.~\eqref{eq:full-objective} can create a conflict that hinders performance. This conflict manifests in two primary ways: 

First, the imbalance between reward signals penalizes exploration. The diversity objective provides a dense and immediate reward based on the discriminator $q_\phi$, which is trained on the states the agent has already visited. In contrast, task rewards are often sparse and delayed (e.g., only given upon success). When the agent attempts to explore a new state necessary to solve the task, the diversity reward drops significantly because the discriminator cannot effectively distinguish latent patterns in unseen regions. Since the sparse task signal cannot immediately compensate for this loss, the agent perceives exploration as a net penalty, incentivizing it to remain in familiar states. 

Second, blindly maximizing diversity distracts the policy from the task goal. The agent may learn to perform distinct but useless behaviors (e.g., reaching for different, but all incorrect, target locations) simply to increase the diversity reward. These pathologies suggest that we should decouple these objectives, enforcing diversity only as a conditional requirement along valid, successful paths.



\subsection{A Decoupled Objective for Exploration-Friendly Diversity}
\label{sec:s_star}
Our core insight is to break the pathological feedback loop by decoupling the diversity objective from the policy's own exploration process. We posit that \emph{diversity should not be sought across all trajectories, which penalizes exploration, but only among those that have already achieved task success}. This ensures the diversity objective encourages meaningful variation within the successful region of the state space, rather than hindering the initial discovery of it.

Before formulation, we define the set of all successful trajectories as
\[
\mathcal{T}_{\text{succ}} = \{ \tau \mid \mathbb{I}_{\text{succ}}(\tau) = 1 \}
\]
where $\mathbb{I}_{\text{succ}}(\tau) \in \{0,1\}$ indicates whether agent completes the task successfully in the trajectory $\tau$,
and then we define $\mathcal{S}^\star$ as the set of all states covered by these successful trajectories:
\[
\mathcal{S}^\star = \bigcup_{\tau=(s_0, a_0, \dots) \in \mathcal{T}_{\text{succ}}} \{s_0, s_1, \dots \} \subseteq \mathcal{S}.
\]
We can then define the state visitation distribution over this manifold, which serves as our target distribution, $d^*(s)$:
\begin{equation}
d^*(s) \triangleq p(s \mid \mathbb{I}_{\text{succ}}(\tau)=1).
\end{equation}
Our ideal objective is to maximize task success and pattern diversity, subject to the constraint that the policy remains on the:
\begin{align}
\label{eq:constrained-objective}
\max_{\theta, \phi} \quad & \mathbb{E}_{z,\tau \sim \pi_\theta(\cdot~|~z)} [R(\tau)] + \beta\, I_\theta(Z; \mathcal{S}^\star) \\
\text{s.t.} \quad & \{ s \in \mathcal{S} \mid d^{\pi_\theta}(s) > 0 \} \subseteq \mathcal{S}^\star
\nonumber
\end{align}
where $ \{s \in \mathcal{S} \mid d^\pi(s) > 0\}$ denotes the set of states 
visited by the policy. The constraint ensures that the policy generates successful trajectories. To make this tractable, we relax the hard constraint into a soft penalty, which leads to the full, unconstrained objective:
\begin{align}
\label{eq:relaxed-objective}
\max_{\theta, \phi} \quad
& \underbrace{
    \mathbb{E}_{z,\tau\sim\pi_\theta(\cdot~|~z)}\big[ R(\tau) \big]
}_{\textcolor{myred}{\textbf{Reinforce}}}
+\, \beta\,
\underbrace{
    I_\theta(Z;\mathcal{S}^\star)
}_{\textcolor{myblue}{\textbf{Discover}}}
\nonumber \\
&\hspace{3.7em}
-\, \alpha\,
\underbrace{
    D\big( d^\pi(s) \, \| \, d^*(s) \big)
}_{\textcolor{mygreen}{\textbf{Learn}}}
\end{align}

where the divergence term $D(\cdot\|\cdot)$ now acts as a soft penalty encouraging the policy to stay on the successful-state manifold. Furthermore, this objective is still intractable to optimize directly because the manifold $\mathcal{S}^\star$ is unknown. This motivates our practical three-stage algorithm designed to approximate this ideal objective.
\subsection{Practical Implementation: A Three-Stage Framework}
\label{sec:three_stage_framework}
The primary challenge in optimizing Eq.~\eqref{eq:relaxed-objective} is that the successful-state manifold $\mathcal{S}^\star$ is unknown. Estimating it online with the learning policy would reintroduce the very exploration bias we aim to eliminate. Our solution is to use a fixed, high-quality dataset of human demonstrations, $\mathcal{D}_{\mathrm{human}}$, as a fixed proxy for the successful-state manifold. This provides a stable target for our learning objectives, decoupling them from the policy's ongoing exploration. This insight motivates our practical three-stage algorithm, which is designed to approximate each of the three terms in our principled objective.

\textbf{Stage 1: \textcolor{myblue}{\textbf{Discover}}.}
To discover meaningful latent patterns from the human data $\mathcal{D}_{\mathrm{human}}$, we adopt a VAE-based framework \cite{vae,vae-clustering,gmm} to maximize the mutual information $I(Z;S)$. This approach trains an encoder, $q_\phi(z|s)$, to map states from successful human trajectories to a latent pattern variable $z$. By learning to reconstruct the states from these latent codes, the encoder is trained to capture the most salient features of the trajectories, effectively clustering them into distinct, semantically meaningful patterns. After training, we refer to these latent codes as the discovered patterns z, and the trained encoder $q_\phi(z|s)$ serves as a fixed model to infer these patterns for the subsequent stages.

\textbf{Stage 2: \textcolor{mygreen}{\textbf{Learn}}}.
After the discovery stage, we use the trained encoder $q_\phi$ to create a labeled dataset, $\mathcal{\widetilde{D}}_{\mathrm{human}}$, 
by assigning a latent $z = \text{argmax}_{z'} q_\phi(z'|s)$ to each state in $\mathcal{D}_{\mathrm{human}}$. We then train a conditional policy $\pi_\theta(a~|~s,z)$ via behavior cloning (BC) on this new dataset:
\begin{equation}\label{eq:behavior-cloning}
\max_\theta\; \mathbb{E}_{(s,a,z)\sim \mathcal{\widetilde{D}}_{\mathrm{human}}}\big[\log \pi_\theta(a\mid s,z)\big].
\end{equation}
The state distribution of the human data, $d_{\mathcal{D}}(s)$, serves as our empirical estimate for the unknown successful-state distribution $d^*(s)$. It is a known result that behavior cloning implicitly minimizes the KL divergence between the policy's state distribution and the expert's state distribution \cite{GAIL}. Therefore, by performing BC on $\mathcal{\widetilde{D}}_{\mathrm{human}}$, we are effectively minimizing $D(d^\pi(s) \| d_{\mathcal{D}}(s))$, thus aligning the initial policy with our best estimate of the successful-state manifold.

\textbf{Stage 3: \textcolor{myred}{\textbf{Reinforce}}.}
In the final stage, we fine-tune the policy using only reward signal from environment, with no MI-based dense shaping required. Since BC only aligns the policy with an incomplete, empirical estimate of the successful manifold, we add a final reinforcement learning stage. For this online stage, we sample patterns $z$ from a fixed distribution such as uniform distribution. By optimizing the task reward, each pattern-conditioned policy is fine-tuned to converge within the true successful manifold $\mathcal{S}^\star$:
\begin{equation}\label{eq:practical-final}
\max_\theta\; \mathbb{E}_{z\sim p(z),\ \tau\sim\pi_\theta(\cdot\mid z)}\big[
R(\tau)\big].
\end{equation}
It is worth noting that although this framework follows the general offline-to-online RL paradigm~\cite{awac,Calql}, but the underlying optimization landscape differs significantly. Figure~\ref{fig:offline_to_online_rl} visually compares the our algorithm against standard baselines.

\subsection{Theoretical Analysis for Diversity Preservation}
\label{sec:theoretical_analysis}
We analyze Stage~3 where each $\pi_\theta(a\mid s,z)$ is refined by PPO using only the sparse, trajectory-level reward $R(\tau)=\gamma^{T-1}\mathbb{I}_{\text{succ}}(\tau)$.
Stages~1--2 provide $K$ patterns $\{z_j\}_{j=1}^K$ and a partition of successful trajectories $\{\mathcal{T}_j^+\}_{j=1}^K$; let $\mathcal{T}_0:=\mathcal{T}\setminus\cup_j\mathcal{T}_j^+$, so $R(\tau)=0$ on $\mathcal{T}_0$.
For pattern $z_j$, we define the \emph{cross-pattern leakage}
\begin{equation}
p_{\mathrm{leak}}^{(j)}(\theta):=\Pr_{\tau\sim p_\theta^j}\!\big[\tau\in\cup_{k\neq j}\mathcal{T}_k^+\big],
\end{equation}
where $p_\theta^j(\tau)=\rho_0(s_0)\prod_{t=0}^{T-1}\pi_\theta(a_t\mid s_t,z_j)\,\mathcal{P}(s_{t+1}\mid s_t,a_t).$
We then give three core assumptions:
(i) (\emph{Failure moat}) Any path between $\mathcal{T}_j^+$ and $\mathcal{T}_k^+$ crosses $\mathcal{T}_0$;
(ii) (\emph{Separated init}) $p_{\mathrm{leak}}^{(j)}(\theta_0)\le\delta_0$;
(iii) (\emph{Proximal PPO target-KL clipping}) $D_{\mathrm{KL}}(p_{\theta_t}^j\|p_{\theta_0}^j)\le E$ for all $t$, with $\pi_{\theta_0}$ initialized after stage 2;
(iv) (\emph{Regularity}) the trajectory score has bounded second moment: $\mathbb{E}[\|\nabla_\theta\log p_\theta^j(\tau)\|^2]\le B$.
\begin{theorem}[Pattern preservation with failure moat and KL-to-init]
\label{thm:mode_preserve}
Under assumptions (i)--(iv), for every pattern $j$ and all Stage~3 iterates $t$,
\[
\Big\|\ \mathbb{E}_{\tau\in\cup_{k\neq j}\mathcal{T}_k^+}\!\big[\nabla_\theta\log p_{\theta_t}^j(\tau)\,R(\tau)\big]\ \Big\|
\ \le\ \sqrt{B}\,\sqrt{\delta_0+\sqrt{E/2}}.
\]
Since $R(\tau)=0$ on $\mathcal{T}_0$, the expected ascent direction is dominated by $\mathcal{T}_j^+$, so PPO updates remain localized and converge to a local optimum within $\mathcal{T}_j^+$.
\end{theorem}
Intuitively, when updating the policy for a given pattern $z_j$, each PPO step aggregates gradients from three types of trajectories:
(i) failures in $\mathcal{T}_0$,
(ii) successful trajectories in $\mathcal{T}_j^+$,
and (iii) successful trajectories belonging to other patterns $\mathcal{T}_k^+$, $k\neq j$.
Type-(i) trajectories have $R(\tau)=0$, so they do not provide positive learning signal and the policy improvement mechanism tends to move probability mass away from such incorrect behaviors.
Type-(iii) trajectories could in principle drag $\pi_\theta(\cdot\mid z_j)$ toward other patterns, but Theorem~\ref{thm:mode_preserve} shows that their total contribution is upper-bounded by a term that scales with $\mathcal{O}(\sqrt{\delta_0+\sqrt{E/2}})$.
When Stage~2 has learned well-separated behaviors (small $\delta_0$) and Stage~3 enforces a small budget $E$, this cross-pattern influence remains weak.
As a result, the overall update direction refines that pattern instead of collapsing different patterns into one. Figure~\ref{fig:three_stage_framework} provides an illustration for understanding.
\section{Experimental Evaluation}
\label{sec:experiments}
\subsection{Experimental Setup}
\label{sec:exp_setup}
\begin{figure}
    \centering
    \includegraphics[width=\linewidth]{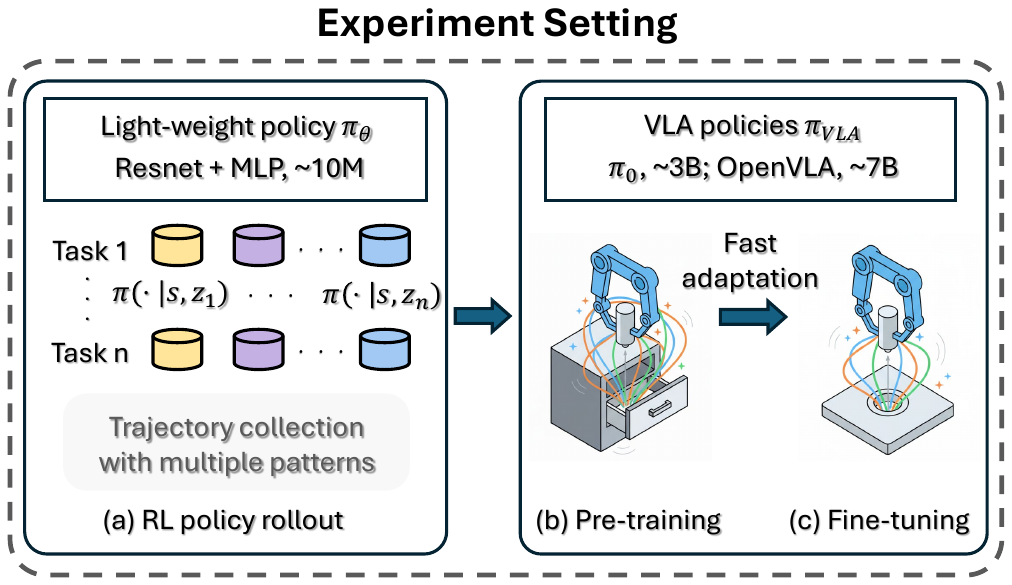}
    \caption{\textbf{Environment Setting and Data Generation Pipeline.} 
    (a) For each task, we train a lightweight RL policy using our DLR framework, then collect high-quality trajectories via policy rollouts. Each color-coded database represents trajectories associated with a distinct behavioral pattern discovered by DLR. We combine data from all tasks for VLA pretraining. 
    (b, c) We use SFT to pretrain variants of VLA architectures on the RL-generated data, then employ the pretrained VLA models for fast adaptation to unseen downstream tasks that the models have never encountered during pretraining.}
    \label{fig:experiment setting}
\end{figure}
We design experiments to answer the following questions:
1)~Does the Discover, Learn, and Reinforce (DLR) framework generate more diverse and successful trajectories compared to standard RL-based data generation methods?
2) Does the increased diversity of the data generated by DLR improve the performance of VLA on downstream tasks?
\begin{figure*}[t]
    \centering
    \includegraphics[width=\textwidth]{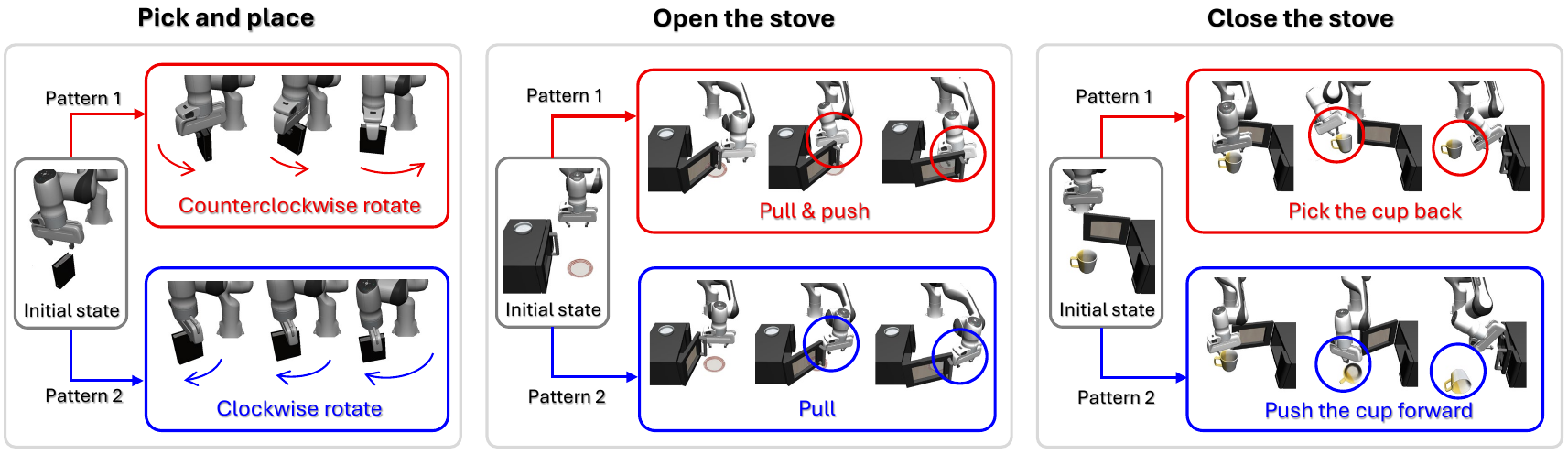}
    \caption{\textbf{Qualitative visualizations of different behaviors discovered in manipulation tasks.} We show that DLR is able to explore diverse behavior patterns under the same initial state. \textbf{Left:} \textit{Pick up the book and place it into the caddy.} The robot needs to adjust the book’s orientation to fit into the upright caddy. DLR learns two strategies: rotating the book clockwise or counterclockwise to align it properly. \textbf{Middle:} \textit{Open the stove.} The robot learns to open the door using different contact strategies, including pulling with the effector's edge followed by a center push, versus consistently pulling with the end-effector. \textbf{Right:} \textit{Close the stove.} A cup blocks the door, requiring obstacle-aware coordination. The robot learns two distinct behaviors, either picking up the cup and placing it backward before closing, or pushing the cup forward to clear the path before closing the stove.}
    \label{fig:skill_visualization}
\end{figure*}
Our experiments are conducted on the LIBERO benchmark~\cite{liu2023libero}, which is explicitly designed to evaluate knowledge transfer for lifelong robot learning. Figure~\ref{fig:experiment setting} illustrates our experimental setup and data generation pipeline.
This makes it an ideal testbed for our hypothesis that diversity in pre-training data improves downstream generalization. We use the \textbf{LIBERO-90} suite as the data source to pre-train the VLA models, which consists of 90 diverse manipulation tasks involving a variety of objects and environments. 
For downstream evaluation, we fine-tune and test the pre-trained models on four other
suites, each designed to probe a different aspect of knowledge transfer:
\begin{itemize}
    \item \textbf{LIBERO-Long}: A set of 10 tasks that are combinations of tasks appearing in LIBERO-90 to evaluate the knowledge transfer to long-horizon tasks.
    \item \textbf{LIBERO-Spatial}: 
    A set of 10 tasks where the spatial relationships among the objects are changed from LIBERO-90 to evaluate generalization on spatial location.
    \item \textbf{LIBERO-Object}: 
    A set of 10 tasks where the objects are changed from LIBERO-90 to evaluate generalization across objects.
    \item \textbf{LIBERO-Goal}: 
    A set of tasks where the goals are changed from LIBERO-90 to evaluate generalization across goals.
\end{itemize}
See detailed description of these tasks and method implementation in Appendix~\ref{app:experiment_details_libero}.
Following previous 
work on generating data using RL~\cite{luo2025precise,rldg,yang2025beyond},
we compare DLR against a strong RL baseline, which is a vanilla offline-to-online RL method initialized via behavior cloning (offline) and then fine-tuned online with proximal policy optimization (PPO) \cite{ppo}, we call it \textbf{O2O-RL}. This represents the conventional single-pattern RL approach. We do not include skill‑discovery baselines, as they optimize only intrinsic‑motivation objectives and typically achieve lower task success, resulting in lower‑quality data. We use pre-trained
Resnet18~\cite{resnet18} to encode the third-person camera view following a MLP head to output executable motion action. For DLR,
we train a conditional policy $\pi(a|s,z)$ using the three-stage pipeline as described in Sec.~\ref{sec:method_framework}, and
use the same environment settings and offline human data source as O2O-RL during training.
The RL algorithm is also PPO. When sampling latent pattern. we set $|Z| = 3$ and use uniform distribution.

We first use different
RL data generation methods to produce RL data on each task in
\textbf{LIBERO-90}, then combine these data to pre-train \(\pi_0\)~\cite{pi0} and OpenVLA~\cite{kim2025openvla} on corresponding datasets.
We keep the data of
roughly the same size and the
same pre-training steps for fair comparison. 
We then fine-tune the pre-trained models on the downstream LIBERO suites for fewer than 3 epochs using the downstream human data from LIBERO-Long, LIBERO-Spatial, LIBERO-Object, and LIBERO-Goal separately. We also compare against a VLA model from VLM checkpoint without pre-training on LIBERO-90.
\begin{figure}[t]
    \centering
    \begin{subfigure}[t]{0.48\linewidth}
        \centering
    \includegraphics[width=\linewidth]{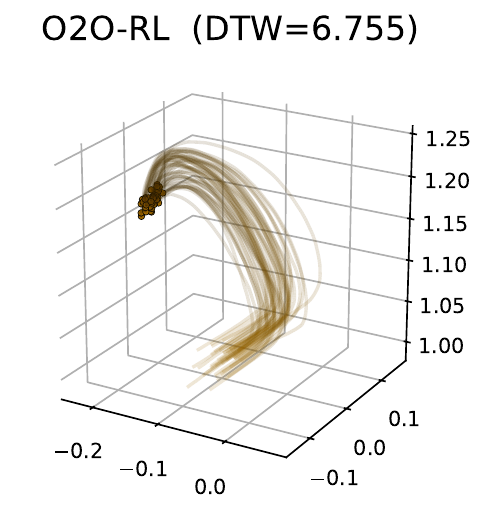}
        \caption{Single-Pattern RL}
    \end{subfigure}
    \hfill
    \begin{subfigure}[t]{0.48\linewidth}
        \centering
        \includegraphics[width=\linewidth]{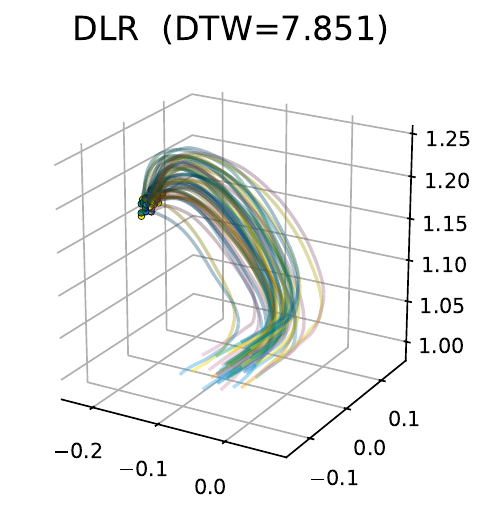}
        \caption{Multi-Pattern (DLR)}
    \end{subfigure}
    \caption{\textbf{Trajectory visualizations.} The task is to close the bottom drawer of the cabinet; the dot indicates the initial position. (a) Standard single-pattern RL converges to a single dominant strategy with limited variation. (b) DLR discovers trajectories with higher variance, DTW refers to the Dynamic Time Warping distance between the two trajectories.}
    \label{fig:traj_compare}
\end{figure}

\subsection{DLR Generates More Diverse and Distinct Patterns than Standard RL}
\begin{figure}[t]
    \centering
    \includegraphics[width=0.80\linewidth]{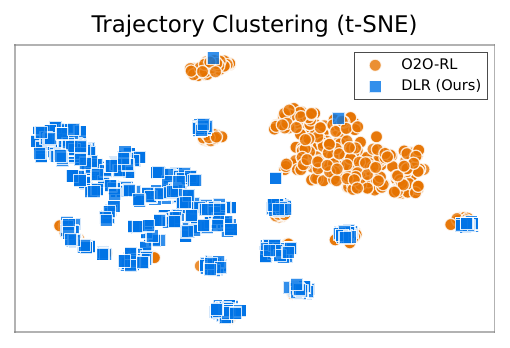}
    \caption{\textbf{t-SNE visualization of trajectory embeddings.} Our DLR-generated trajectories (blue squares) show broad, multi-modal coverage of the embedding space. In contrast, the single-pattern RL baseline (orange circles) exhibits severe mode collapse, with trajectories concentrating in a few dense clusters, demonstrating a lack of diversity.}
    \label{fig:tsne_clustering}
\end{figure}

First, we verify that DLR produces a richer and more structured data distribution compared to the standard O2O-RL baseline. 
We provide qualitative evidence of this increased diversity in Figures~\ref{fig:skill_visualization}, \ref{fig:traj_compare} and \ref{fig:tsne_clustering}, and quantitative support in
Table~\ref{tab:diversity_metrics} These results highlight a key distinction: while the O2O-RL baseline tends to collapse to a single deterministic solution, DLR successfully recovers multiple distinct and valid behavioral patterns.

\begin{table}[t]
    \centering
    \caption{\textbf{Trajectory Diversity Metrics.} DLR consistently produces trajectories with higher diversity across all metrics compared to the single-pattern RL baseline.}
    \label{tab:diversity_metrics}
    \begin{tabular}{lcc}
        \toprule
        \textbf{Metric} & \textbf{O2O-RL} & \textbf{DLR} \\
        \midrule
        Mean Pairwise Distance ($\uparrow$) & 10.487 & \textbf{26.405} \\
        Endpoint Variance ($\uparrow$)      & 0.092  & \textbf{1.170}  \\
        Direction Variance ($\uparrow$)     & 0.083  & \textbf{0.085}  \\
        Path Length Variance ($\uparrow$)   & 17.193 & \textbf{26.546} \\
        \bottomrule
    \end{tabular}
\end{table}

\begin{table}[t]
    \centering
    \caption{\textbf{Downstream Task Adaptation.} Success rates (\%) on fast adapted downstream tasks for both \(\pi_0\) and OpenVLA models. We run 50 training runs for each task and report the average success rate.}
    \label{tab:adaptation_downstream_combined}
    \setlength{\tabcolsep}{4pt}
    \begin{tabular}{lccccc}
        \toprule
        \textbf{Data Source} & \textbf{Spatial} & \textbf{Object} & \textbf{Goal} & \textbf{Long} & \textbf{Avg.} \\
        \midrule
        \multicolumn{6}{l}{\textbf{Model: \(\pi_0\)}} \\
        No Pretraining & 0.60 & 0.00 & 0.20 & 0.00 & 0.20 \\
        O2O-RL & 4.40 & 15.20 & 19.20 & 15.00 & 13.45 \\
        \textbf{DLR (Ours)} & \textbf{6.00} & \textbf{18.60} & \textbf{22.40} & \textbf{18.80} & \textbf{16.45} \\
        \midrule
        \multicolumn{6}{l}{\textbf{Model: OpenVLA}} \\
        No Pretraining & 1.60 & 2.60 & 17.04 & 2.20 & 5.86 \\
        O2O-RL & 11.60 & 3.04 & 19.11 & 3.94 & 9.42 \\
        \textbf{DLR (Ours)} &\textbf{ 19.88} & \textbf{3.67} & \textbf{34.20} & \textbf{9.21} & \textbf{16.72} \\
        \bottomrule
    \end{tabular}
\end{table}
We visualize the generated behaviors to highlight the diversity in behaviors in Figure~\ref{fig:traj_compare}. We observe that standard single-pattern RL converges to a single dominant strategy with limited variation. 
In contrast, DLR discovers multiple distinct strategies (shown in different colors), covering a much wider area of the state space. We also plot the t-SNE embedding in Figure~\ref{fig:tsne_clustering} for further analysis.
We find that the trajectories of the baseline exhibit a clear mode collapse into a single dominant cluster, a behavior predicted by theory for naive reward maximization. In contrast, our DLR-generated trajectories are spread widely across the space in multiple distinct clusters, confirming that our method generates a substantially more diverse and structured dataset.

Beyond visual inspection, we quantify trajectory diversity using the metrics in Table~\ref{tab:diversity_metrics}.
Across all these metrics, DLR consistently demonstrates improved diversity compared to the baseline. See Appendix~\ref{app:additional_experiments} for detailed analysis and definition of the metrics.

\subsection{Data Diversity Translates to Superior Downstream Performance}

\begin{figure}[t]
    \centering
    \includegraphics[width=0.8\linewidth]{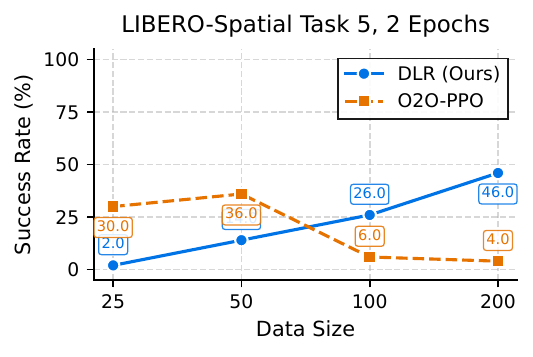}
    \caption{We observe a positive scaling trend. The x-axis refers to the amount of RL-generated data for each task (data size * 90 tasks = total data size), and the y-axis refers to the success rate of \(\pi_0\) on the downstream task.}
    \label{fig:scaling_curve}
\end{figure}

To investigate whether the enhanced diversity in the data generated by DLR leads to better downstream VLA performance, we report the success rate of different models fine-tuned on the downstream tasks in Table~\ref{tab:adaptation_downstream_combined}.
\(\pi_0\) and OpenVLA models. We observe that the models pre-trained on the data generated by DLR consistently outperform those trained on the data from O2O-RL and the VLM baseline. The performance gains are particularly pronounced on the out-of-distribution generalization suites, indicating that the pre-training dataset containing more diverse behaviors enables better skill transfer. To
analyze the effect of data quantity, we show the scaling curve of downstream performance w.r.t.
the amount of RL-generated data (holding human data fixed) in Figure~\ref{fig:scaling_curve}. We hypothesize a positive scaling trend, demonstrating the value of our algorithmic data generation pipeline.
\section{Conclusion}
\label{sec:conclusion}In this paper, we propose a three-stage framework, DLR, to encourage behavioral diversity in data generated for VLA pre-training. 
Theoretically, we show how our decoupled offline discovery and online reinforcement learning design avoids the exploration conflict that leads to mode collapse in naive diversity-maximization approaches, and we provide a theoretical analysis showing that our method preserves pattern diversity during RL. 
Empirically, our experiments on the LIBERO benchmark demonstrate that DLR generates a more diverse dataset than a strong RL baseline, and that this diversity translates into better downstream generalization on unseen tasks. 
In future work, we will explore combining DLR with automated task and environment generation to create a fully autonomous pipeline capable of systematically scaling up data generation for embodied foundation models.

{
    \small
    \bibliographystyle{ieeenat_fullname}
    \bibliography{main}
}

\clearpage

\setcounter{section}{0}
\renewcommand{\thesection}{\Alph{section}}

\setcounter{page}{1}
\maketitlesupplementary

\section{Mutual Information Objectives and Exploration Conflict}

In Sec.~\ref{sec:general_infomax} we combine task reward with a mutual-information-based diversity objective.
This section provides detailed derivations and a more formal analysis of why such objectives can create an
exploration conflict, for both the reverse and forward variational bounds.

\subsection{Reverse Variational Bound}
\label{app:mi_derivation}
For two random variables $(Z,Y)$ with joint density $p(z,y)$,
the mutual information can be written as
\begin{equation}
I(Z;Y) = \mathbb{E}_{p(z,y)}[\log p(z~|~y) - \log p(z)].
\end{equation}
Introducing any variational distribution $q_\phi(z~|~y)$ and multiplying and dividing by it inside the logarithm yields
\begin{align}
I(Z;Y)
&= \mathbb{E}_{p(z,y)}\!\left[\log \frac{p(z|y)}{p(z)}\right] \nonumber \\
&= \mathbb{E}_{p(z,y)}\!\left[\log \frac{q_\phi(z|y)}{p(z)}
      + \log \frac{p(z|y)}{q_\phi(z|y)}\right] \nonumber \\
&= \mathbb{E}_{p(z,y)}[\log q_\phi(z|y) - \log p(z)]
   + \mathbb{E}_{p(y)}\mathrm{KL}\!\big(p(z|y)\,\|\,q_\phi(z|y)\big).
\label{eq:mi_lower_bound_derivation_app}
\end{align}
Since the KL divergence is non-negative, we obtain the reverse variational bound
\begin{equation}
I(Z;Y) \;\ge\; \mathbb{E}_{p(z,y)}[\log q_\phi(z|y) - \log p(z)].
\end{equation}

Setting $Y=S$ and $p(z,s)=p_\theta(z,s)$ recovers the bound used in Sec.~\ref{sec:general_infomax}:
\begin{equation}
I_\theta(Z;S)
\;\ge\;
\mathbb{E}_{z \sim p(z),\, s \sim d^{\pi_\theta(\cdot|\cdot,z)}}[\log q_\phi(z~|~s) - \log p(z)],
\end{equation}
which naturally defines the intrinsic diversity reward
\begin{equation}
    \label{eq:intrinsic-reward-app}
    r_{\mathrm{div}}(s,z) \;\triangleq\; \log q_\phi(z~|~s) - \log p(z).
\end{equation}
Optimizing the full objective in Eq.~\eqref{eq:full-objective} is therefore equivalent to maximizing
task return plus this variational lower bound on $I_\theta(Z;S)$.

\subsection{Proof of the Exploration-Penalty Proposition}
\label{app:proof_exploration_cost}

We now justify the claim in Sec.~\ref{sec:pathology} that the MI-based intrinsic reward
can penalize exploration.
Recall that the discriminator $q_\phi(z|s)$ is trained \emph{on-policy}, so it learns the empirical
posterior over patterns under the current policy:
\begin{equation}
q_\phi(z|s) \approx d^\pi(z|s).
\end{equation}
Under this approximation, the intrinsic reward in Eq.~\eqref{eq:intrinsic-reward} satisfies
\begin{equation}
r_{\mathrm{div}}(s,z) \approx \log d^\pi(z|s) - \log p(z).
\end{equation}

\noindent\textbf{Unseen states.}
For any state $s$ that has not yet been visited by the policy, the discriminator receives no data and
cannot specialize across patterns. A natural and standard behavior is that its prediction reverts
to the prior over patterns:
\begin{equation}
d^\pi(z|s) \approx p(z) \quad\Longrightarrow\quad
r_{\mathrm{div}}(s,z) \approx \log p(z) - \log p(z) = 0,
\end{equation}
which means estimator does not know which pattern the unseen belong to and system are no any information for unseen state.
\noindent\textbf{Highly discriminable familiar states.}
Conversely, consider a state $s'$ that is visited frequently and is highly informative about a
particular pattern $z'$. In this case the empirical posterior becomes sharply peaked,
$d^\pi(z'|s') \approx 1$ and $d^\pi(z|s') \approx 0$ for $z\neq z'$. The intrinsic reward for $(s',z')$ is
\begin{equation}
r_{\mathrm{div}}(s',z') \approx \log 1 - \log p(z') = -\log p(z').
\end{equation}
Under a uniform prior over $K$ patterns, $p(z)=1/K$ and the maximum reward is $\log K$.

\noindent\textbf{Consequence for the combined objective.}
In the full objective
\begin{equation}
    J(\theta)
    = \mathbb{E}_{z,\tau,s}\!\left[ R(\tau) + \beta\, r_{\mathrm{div}}(s,z) \right]
    \label{eq:full-objective-app}
\end{equation}
the task reward $R(\tau)$ is sparse and delayed, whereas $r_{\mathrm{div}}(s,z)$ is dense and depends only
on the current state and discriminator predictions.
Any policy update that moves probability mass from a familiar, highly discriminable state $s'$ to an unseen state $s$
immediately replaces a positive bonus
$\beta\,\log K$ by approximately zero intrinsic reward, while the task return $R(\tau)$ is unchanged in
the short term.
For sufficiently large $\beta$, the instantaneous change in $J(\theta)$ is therefore negative, so gradient-based
updates are biased \emph{against} such exploratory moves.
As a result, the policy is encouraged to remain in a narrow set of familiar states that are easy for
the discriminator, rather than exploring new regions where higher task rewards may lie.

\subsection{Forward Variational Bound and Additional Conflict Modes}
\label{app:forward_bound}

For completeness, we also consider the \emph{forward} variational bound, which rewrites the same mutual
information as
\begin{equation}
I(Z;S) = \mathbb{E}_{p(z,s)}[\log p(s~|~z) - \log p(s)].
\end{equation}
Introducing a variational decoder $q_\phi(s|z)$ and repeating the argument yields
\begin{align}
I(Z;S)
&= \mathbb{E}_{p(z,s)}\!\left[\log \frac{p(s|z)}{p(s)}\right] \nonumber \\
&= \mathbb{E}_{p(z,s)}\!\left[\log \frac{q_\phi(s|z)}{p(s)}
      + \log \frac{p(s|z)}{q_\phi(s|z)}\right] \nonumber \\
&\ge \mathbb{E}_{p(z,s)}[\log q_\phi(s|z) - \log p(s)].
\end{align}
This leads to the forward intrinsic reward
\begin{equation}
    r_{\text{fwd}}(s,z) = \log q_\phi(s~|~z) - \log p(s),
\end{equation}
which is the form discussed in Sec.~\ref{sec:pathology} when $p(s)$ is instantiated as the
state visitation distribution $\rho_\pi(s)$ of the current policy.

Compared to $r_{\mathrm{div}}$, the forward reward favors states that are both
\emph{predictable} under the decoder and relatively \emph{unlikely} under the current policy:
\begin{equation}
r_{\text{fwd}}(s,z)
  = \log \frac{q_\phi(s|z)}{\rho_\pi(s)}.
\end{equation}
Because $q_\phi(s|z)$ is trained only on states that the policy has already visited,
it is a poor model of truly novel states. The ratio $q_\phi(s|z)/\rho_\pi(s)$ is therefore maximized
not at genuinely unexplored regions, but at states that are minor, predictable perturbations of
familiar ones: $\rho_\pi(s)$ is small enough to yield ``novelty'', yet $q_\phi(s|z)$ remains large because
the decoder has seen similar states.

Thus, the forward MI surrogate encourages the policy to seek ``novel but easy'' states close to its
current support, rather than to explore genuinely unseen parts of the state space.
Together with the reverse-bound analysis above, this yields a unified picture:
both on-policy MI surrogates couple the diversity objective tightly to the \emph{current} visitation
distribution, which systematically biases the agent toward practicing what it already knows instead of
discovering new, task-relevant behaviors.
\clearpage
\section{Proof of Theorem~\ref{thm:mode_preserve}}
\paragraph{Notation recap.}
For a fixed pattern $z_j$, write $p_\theta^j(\tau):=p_\theta(\tau\mid z_j)$ for the trajectory distribution induced by $\pi_\theta(a\mid s,z_j)$.
Let the disjoint successful components be $\{\mathcal{T}_j^+\}_{j=1}^K$ and the failure set be $\mathcal{T}_0:=\mathcal{T}\setminus\cup_{j=1}^K\mathcal{T}_j^+$, on which $R(\tau)=\gamma^{T(\tau)-1}\,\mathbb{I}_{\mathrm{succ}}(\tau)=0$.
Define the cross-pattern event $A_j:=\cup_{k\neq j}\mathcal{T}_k^+$ and the leakage
$p_{\mathrm{leak}}^{(j)}(\theta):=\Pr_{\tau\sim p_\theta^j}[\,\tau\in A_j\,]$.
Assume the KL-to-init budget $D_{\mathrm{KL}}(p_{\theta_t}^j\|p_{\theta_0}^j)\le E$ and the score-moment bound
$\mathbb{E}_{p_\theta^j}\!\big[\|\nabla_\theta\log p_\theta^j(\tau)\|^2\big]\le B$ hold for all iterates visited in Stage~3.

\begin{lemma}[Trajectory-level policy-gradient identity]
\label{lem:pg-trajectory}
For $J_j(\theta):=\mathbb{E}_{\tau\sim p_\theta^j}[R(\tau)]$,
\begin{equation}
\label{eq:traj-pg}
\nabla_\theta J_j(\theta)
~=~\mathbb{E}_{\tau\sim p_\theta^j}\!\big[\nabla_\theta \log p_\theta^j(\tau)\,R(\tau)\big].
\end{equation}
Moreover, $\log p_\theta^j(\tau)=\sum_{t=0}^{T(\tau)-1}\log \pi_\theta(a_t\mid s_t,z_j)$ since environment dynamics do not depend on $\theta$.
\end{lemma}

\begin{proof}
By dominated convergence (bounded $R(\tau)\in[0,1]$) and the log-derivative trick:
$\nabla J_j=\int \nabla p_\theta^j(\tau)\,R(\tau)\,d\tau=\int p_\theta^j(\tau)\,\nabla\log p_\theta^j(\tau)\,R(\tau)\,d\tau$.
The second claim follows because $p_\theta^j(\tau)$ factors into $\rho_0$ and $\mathcal{P}$ (both $\theta$-independent) times $\prod_t \pi_\theta(a_t\mid s_t,z_j)$.
\end{proof}

\begin{lemma}[Pinsker]
\label{lem:pinsker}
For any distributions $P,Q$ on $(\mathcal{T},\mathcal{F})$,
$\mathrm{TV}(P,Q):=\sup_{A\in\mathcal{F}}|P(A)-Q(A)| \le \sqrt{\tfrac{1}{2}D_{\mathrm{KL}}(P\|Q)}$.
\end{lemma}

\begin{lemma}[Leakage growth under a KL-to-init budget]
\label{lem:leakage}
For any iterate $\theta_t$,
\begin{equation}
\label{eq:leakage}
p_{\mathrm{leak}}^{(j)}(\theta_t)
~=~p_{\theta_t}^j(A_j)
~\le~ p_{\theta_0}^j(A_j)+\mathrm{TV}\!\big(p_{\theta_t}^j,p_{\theta_0}^j\big)
~\le~ \delta_0+\sqrt{E/2}.
\end{equation}
\end{lemma}

\begin{proof}
By the definition of total variation distance,
$|p_{\theta_t}^j(A_j)-p_{\theta_0}^j(A_j)|\le \mathrm{TV}(p_{\theta_t}^j,p_{\theta_0}^j)$.
Using $p_{\theta_0}^j(A_j)\le \delta_0$ (separated init) and Lemma~\ref{lem:pinsker} with the KL budget yields \eqref{eq:leakage}.
\end{proof}

\begin{lemma}[Cross-pattern gradient bound]
\label{lem:crossgrad}
Let $g_{\mathrm{cross}}^{(j)}(\theta):=\mathbb{E}_{p_\theta^j}\!\big[\nabla_\theta\log p_\theta^j(\tau)\,R(\tau)\,\mathbf{1}\{\tau\in A_j\}\big]$.
Then
\begin{equation}
\label{eq:crossgrad}
\|g_{\mathrm{cross}}^{(j)}(\theta)\|
~\le~ \sqrt{\mathbb{E}_{p_\theta^j}\!\big[\|\nabla_\theta\log p_\theta^j(\tau)\|^2\big]}\;
\sqrt{\mathbb{E}_{p_\theta^j}\!\big[R(\tau)^2\,\mathbf{1}\{\tau\in A_j\}\big]}
~\le~ \sqrt{B}\,\sqrt{p_{\mathrm{leak}}^{(j)}(\theta)}.
\end{equation}
\end{lemma}

\begin{proof}
Apply Cauchy--Schwarz to $\mathbb{E}[XY]$ with
$X=\nabla_\theta\log p_\theta^j(\tau)$ and $Y=R(\tau)\mathbf{1}\{\tau\in A_j\}$, and use $R(\tau)\in[0,1]$.
\end{proof}

\begin{lemma}[Failure term does not create positive ascent]
\label{lem:failure}
Let $g_{\mathrm{fail}}^{(j)}(\theta):=\mathbb{E}_{p_\theta^j}\!\big[\nabla_\theta\log p_\theta^j(\tau)\,R(\tau)\,\mathbf{1}\{\tau\in \mathcal{T}_0\}\big]$.
Then $g_{\mathrm{fail}}^{(j)}(\theta)=0$ since $R(\tau)=0$ on $\mathcal{T}_0$.
Furthermore, when PPO uses a per-timestep advantage baseline, the additional baseline term has zero expectation and does not alter the result.\footnote{For any function $b_t(s_t)$ independent of $a_t$,
$\mathbb{E}\big[\sum_t\nabla_\theta\log \pi_\theta(a_t\mid s_t,z_j)\,b_t(s_t)\big]
=\mathbb{E}\big[\sum_t b_t(s_t)\,\mathbb{E}_{a_t\sim\pi_\theta(\cdot\mid s_t,z_j)}[\nabla_\theta\log \pi_\theta(a_t\mid s_t,z_j)]\big]=0$.}
\end{lemma}

\begin{proof}
Immediate from $R(\tau)\equiv 0$ on $\mathcal{T}_0$ and the footnote argument for the baseline term.
\end{proof}

\begin{proof}[Proof of Theorem~\ref{thm:mode_preserve}]
For a fixed pattern $z_j$, decompose \eqref{eq:traj-pg} into three regions:
\begin{align*}
\nabla_\theta J_j(\theta_t)
&=\underbrace{\mathbb{E}_{p_{\theta_t}^j}\!\big[\nabla\log p_{\theta_t}^j(\tau)\,R(\tau)\,\mathbf{1}\{\tau\in \mathcal{T}_j^+\}\big]}_{\text{within-pattern}}
+\underbrace{\mathbb{E}_{p_{\theta_t}^j}\!\big[\nabla\log p_{\theta_t}^j(\tau)\,R(\tau)\,\mathbf{1}\{\tau\in A_j\}\big]}_{\text{cross-pattern}}
+\underbrace{\mathbb{E}_{p_{\theta_t}^j}\!\big[\nabla\log p_{\theta_t}^j(\tau)\,R(\tau)\,\mathbf{1}\{\tau\in \mathcal{T}_0\}\big]}_{\text{failure}}.
\end{align*}
By Lemma~\ref{lem:failure} the failure term is zero.
By Lemmas~\ref{lem:leakage} and \ref{lem:crossgrad},
\[
\big\|\text{cross-pattern}\big\|\ \le\ \sqrt{B}\,\sqrt{\,\delta_0+\sqrt{E/2}\,}.
\]
Hence the ascent direction is dominated by successful trajectories in $\mathcal{T}_j^+$,
which keeps PPO updates localized to the initialized component and yields convergence to a stationary point within $\mathcal{T}_j^+$ under standard stochastic-approximation conditions.
\end{proof}

\paragraph{Remark on PPO surrogate.}
PPO implements a clipped surrogate but also monitors a target KL; assuming the per-pattern cumulative KL-to-init constraint
$D_{\mathrm{KL}}(p_{\theta_t}^j\|p_{\theta_0}^j)\le E$ (enforced by early stopping or penalties),
Pinsker converts it to $\mathrm{TV}(p_{\theta_t}^j,p_{\theta_0}^j)\le \sqrt{E/2}$, which is precisely the quantity needed in Lemma~\ref{lem:leakage}.
The baseline/advantage control variate leaves the expected gradient unchanged (Lemma~\ref{lem:failure}, footnote), so the trajectory-level proof carries over.
\clearpage
\section{Additional Experimental Results}
\label{app:additional_experiments}

\subsection{Definition of Trajectory Diversity Metrics}
To quantitatively evaluate the diversity of the generated trajectories, we compute several metrics on the feature embeddings of the trajectories. Let a dataset of $N$ trajectories be represented by their embeddings $\{\tau_1, \tau_2, \dots, \tau_N\}$, where each $\tau_i$ is a sequence of state embeddings.

\paragraph{Mean Pairwise Distance.} This metric measures the average distance between all pairs of trajectories in the embedding space. A higher value indicates greater overall separation. It is defined as:
\[
\text{Dist}_{\text{mean}} = \frac{1}{N(N-1)/2} \sum_{i < j} \|\tau_i - \tau_j\|_2
\]
where we use a feature representation (e.g., from a pre-trained VAE) for each trajectory $\tau_i$.

\paragraph{Endpoint Variance.} This metric captures the diversity of the final outcomes. A higher variance suggests that the trajectories end in a wider variety of states. It is defined as:
\[
\text{Var}_{\text{end}} = \text{Var}(\{s_{T_i}^{(i)} \mid i=1,\dots,N\})
\]
where $s_{T_i}^{(i)}$ is the final state of trajectory $\tau_i$.

\paragraph{Direction Variance.} This metric measures the diversity in the overall direction of motion. It is calculated as the variance of the normalized direction vectors from the start to the end of each trajectory:
\[
\text{Var}_{\text{dir}} = \text{Var}\left(\left\{\frac{s_{T_i}^{(i)} - s_0^{(i)}}{\|s_{T_i}^{(i)} - s_0^{(i)}\|_2}\right\}_{i=1}^N\right)
\]
where $s_0^{(i)}$ and $s_{T_i}^{(i)}$ are the initial and final states of trajectory $\tau_i$.

\paragraph{Path Length Variance.} This metric captures the variation in the total length of the paths taken by the agent. A higher variance indicates that the trajectories follow paths of more varied lengths. It is defined as:
\[
\text{Var}_{\text{len}} = \text{Var}\left(\left\{\sum_{t=0}^{T_i-1} \|s_{t+1}^{(i)} - s_t^{(i)}\|_2\right\}_{i=1}^N\right)
\]

\subsection{Additional Diversity Analysis: Distance Matrices}
\begin{figure}[h]
    \centering
    \begin{subfigure}[b]{0.48\linewidth}
        \centering
        \includegraphics[width=\linewidth]{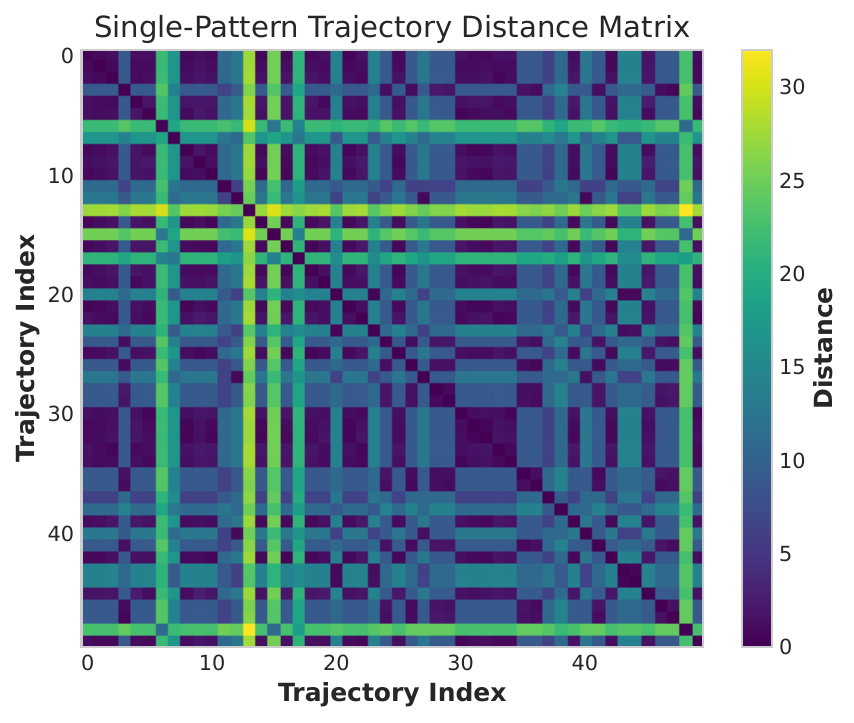}
        \caption{O2O-RL}
    \end{subfigure}
    \hfill
    \begin{subfigure}[b]{0.48\linewidth}
        \centering
        \includegraphics[width=\linewidth]{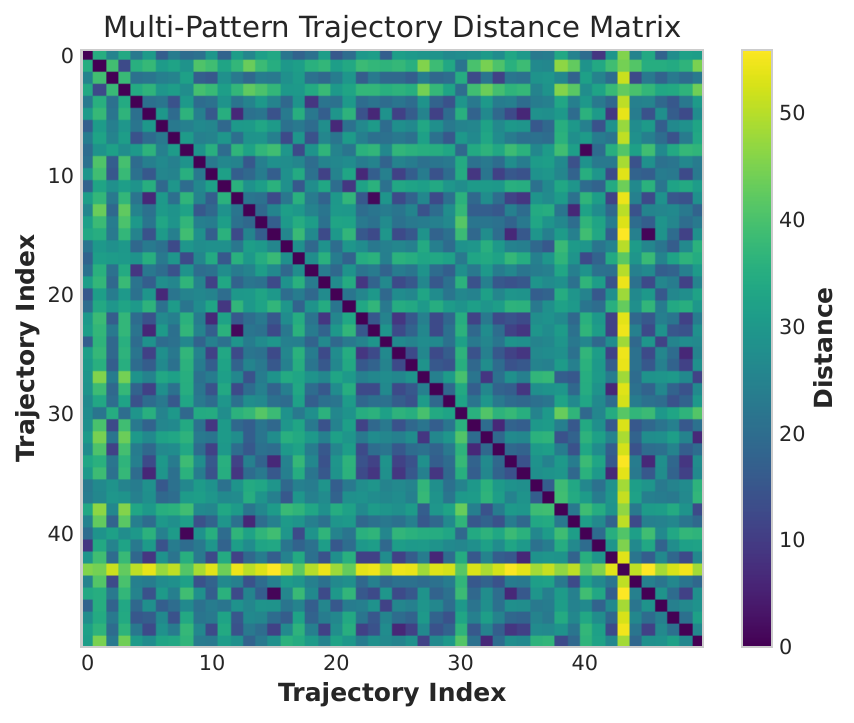}
        \caption{DLR}
    \end{subfigure}
    \caption{\textbf{Distance matrices.} O2O-RL shows relatively uniform distances indicating similar trajectories, while DLR reveals clear block structure indicating distinct behavioral modes.}
    \label{fig:distance_matrices_supp}
\end{figure}

To further visualize the structure of the generated trajectories, we compute and visualize the pairwise distance matrices between trajectory embeddings. As shown in Figure~\ref{fig:distance_matrices_supp}, the matrix for the O2O-RL baseline shows relatively uniform, low distances, indicating all trajectories are highly similar to one another. In contrast, the distance matrix for DLR reveals a clear block-diagonal structure, which provides strong evidence that our method discovers distinct and well-separated behavioral modes.

\begin{figure}[h]
    \centering
    \includegraphics[width=0.7\linewidth]{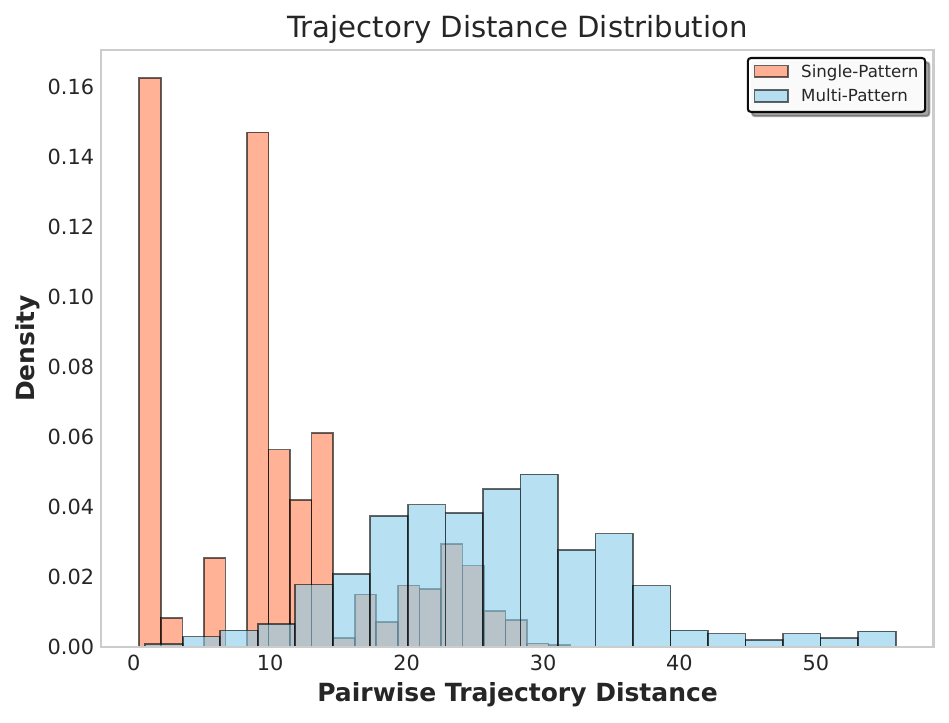}
    \caption{\textbf{Pairwise distance distributions.}}
    \label{fig:pairwise_distance_supp}
\end{figure}

The distribution of these pairwise distances is shown in Figure~\ref{fig:pairwise_distance_supp}. The distribution for DLR is shifted to the right and is more dispersed, indicating a higher average distance and greater variance between trajectories compared to the O2O-RL baseline.

\clearpage
\section{Experiments details on LIBERO}
\label{app:experiment_details_libero}
This section provides additional details of the experimental protocol on the LIBERO benchmark used in Sec.~\ref{sec:experiments}.

Success rate is our primary evaluation metric: for each task we execute 50 rollouts of a policy with different random seeds and report the fraction of trajectories that satisfy the task-specific success predicate provided by the LIBERO environment.

\subsection{Experimental Setup}

We follow the standard LIBERO evaluation protocol~\cite{liu2023libero}.
All experiments are conducted in the simulation environments released with LIBERO, using the default robot and scene configurations.
LIBERO-90 serves as the source suite for data generation and VLA pretraining, while four additional suites---LIBERO-Long, LIBERO-Spatial, LIBERO-Object, and LIBERO-Goal---are used only for downstream fine-tuning and evaluation.
Unless otherwise specified, we use the same observation modalities, control frequency, and sparse binary reward functions as the original benchmark.

\subsection{Task Details}

LIBERO-90 contains 90 diverse tabletop manipulation tasks with natural-language instructions, covering object rearrangement, container manipulation, articulated objects, and obstacle-aware behaviors.
Each task comes with human demonstration trajectories that we treat as successful examples.
The four downstream suites probe different generalization axes relative to LIBERO-90 as summarized in Sec.~\ref{sec:experiments}: task composition (LIBERO-Long), spatial rearrangements (LIBERO-Spatial), object changes (LIBERO-Object), and goal changes (LIBERO-Goal).
For all suites, an episode terminates either when the simulator reports task success or when a fixed horizon is reached, in which case the episode is marked as a failure.

\subsection{Data Generation Details}

For each LIBERO-90 task, we construct two RL-based data generators: the single-pattern O2O-RL baseline and our multi-pattern DLR framework.
Both methods start from the same offline human demonstrations, which we split into training and validation sets.
O2O-RL first trains a single policy by behavior cloning on all demonstrations of a task, and then fine-tunes it online with PPO using only sparse task rewards.

DLR follows the three-stage procedure in Sec.~\ref{sec:method_framework}.
In Stage~1 (\textcolor{myblue}{Discover}), we use a clustering method to encode successful human trajectories into a latent space $\mathcal{Z}$. Instead of encoding high-dimensional trajectories directly, the offline human demonstrations in LIBERO provide proprioceptive states. We train an encoder $q(z|\text{state})$ to discover $K = 3$ latent patterns for each task; this encoder is then used to label the current human trajectories.
In Stage~2 (\textcolor{mygreen}{Learn}), we assign a pattern label $z$ to each state via $\arg\max_{z'} q_\phi(z'|s)$ and train a conditional policy $\pi_\theta(a|s,z)$ by behavior cloning on the labeled dataset.
In Stage~3 (\textcolor{myred}{Reinforce}), we fine-tune $\pi_\theta(a|s,z)$ with PPO under sparse task reward, sampling $z$ from a fixed prior $p(z)$ (uniform in our experiments) and keeping $q_\phi$ frozen.
After convergence, we generate RL trajectories for each task by rolling out the final policies; these trajectories, together with task instructions, form the RL-generated dataset used to pretrain VLA models.
We keep the total number of RL trajectories per task approximately matched between DLR and O2O-RL to ensure a fair comparison of data quality rather than quantity.

\subsection{Model Details}

\paragraph{RL policies.}
For both O2O-RL and DLR, the policy takes as input the third-person RGB observation of the scene and low-dimensional proprioceptive features (e.g., end-effector pose and gripper state).
We follow Sec.~\ref{sec:experiments} and use a ResNet18 backbone~\cite{resnet18} to encode the image into a feature vector, concatenate it with proprioceptive inputs, and pass the result through an MLP head to output continuous actions.
In DLR, the latent code $z$ is embedded and concatenated with the state features before the MLP head.

\paragraph{VLA models.}
For downstream pretraining, we adopt two representative architectures: \(\pi_0\)~\cite{pi0} and OpenVLA~\cite{kim2025openvla}.
Both models take language instructions and visual observations as input and predict low-level actions; we follow the official implementations and hyperparameters of the respective works where possible, and only vary the source of pretraining data (DLR vs.\ O2O-RL vs.\ no RL pretraining).

\subsection{Model Training Details}

\paragraph{RL training.}
We use PPO~\cite{ppo} for all online RL updates, with standard hyperparameters similar to~\cite{rldg,yang2025beyond}.
Human demonstrations are used only in the offline behavior cloning stage (for O2O-RL and Stage~2 of DLR); online fine-tuning relies solely on sparse task rewards.
We monitor task success rate on a held-out validation subset of environments and stop RL training early once performance saturates, to avoid overfitting a single pattern and to respect a fixed compute budget per task.

\paragraph{VLA pretraining.}
Given the RL-generated datasets from each method on LIBERO-90, we pretrain \(\pi_0\) and OpenVLA by supervised fine-tuning (SFT) on trajectory data, minimizing the negative log-likelihood of actions conditioned on the instruction and observations.
We control for the total number of pretraining steps and the number of trajectories, so that differences in downstream performance can be attributed primarily to the structure and diversity of the data rather than scale.

\subsection{Model Fine-tuning Details}

For downstream evaluation, we fine-tune the pretrained VLA models separately on each of the four LIBERO suites using only the human demonstrations provided for those suites.
Following Sec.~\ref{sec:experiments}, we run fewer than three epochs of fine-tuning on each downstream dataset, using early stopping based on validation success rate to mimic a realistic low-data adaptation regime.
After fine-tuning, we evaluate each model for 50 rollouts per task and report the average success rate across tasks within each suite, as summarized in Table~\ref{tab:adaptation_downstream_combined}.

\clearpage
\section{MI-Shaped Objective: Negative Result and Analysis}
\label{app:mi_shaped_negative}

\begin{figure}[h]
    \centering
    \begin{subfigure}[b]{0.48\linewidth}
        \centering
        \includegraphics[width=\linewidth]{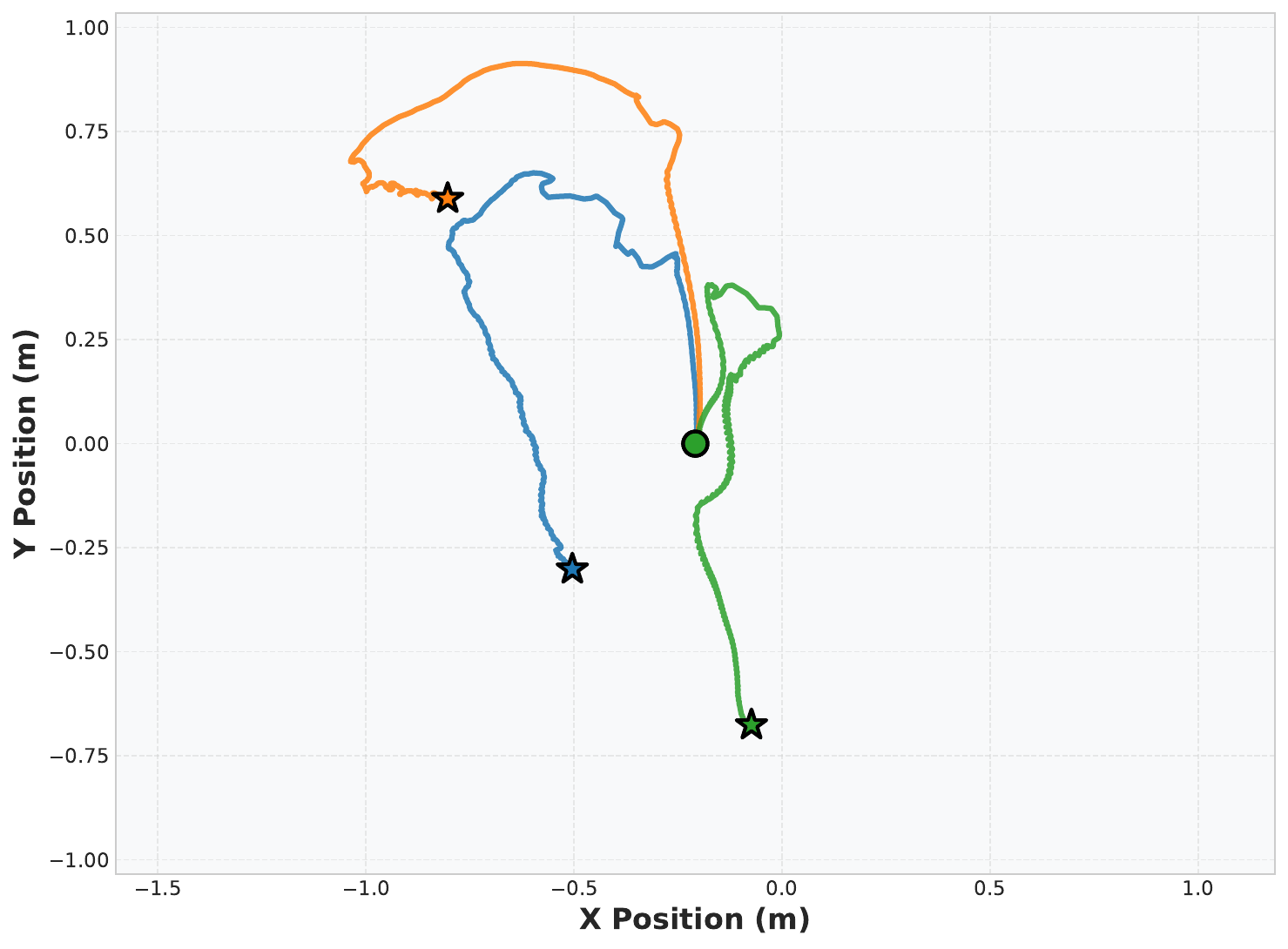}
        \caption{MI-shaped (Eq.~\eqref{eq:full-objective})}
    \end{subfigure}
    \hfill
    \begin{subfigure}[b]{0.48\linewidth}
        \centering
        \includegraphics[width=\linewidth]{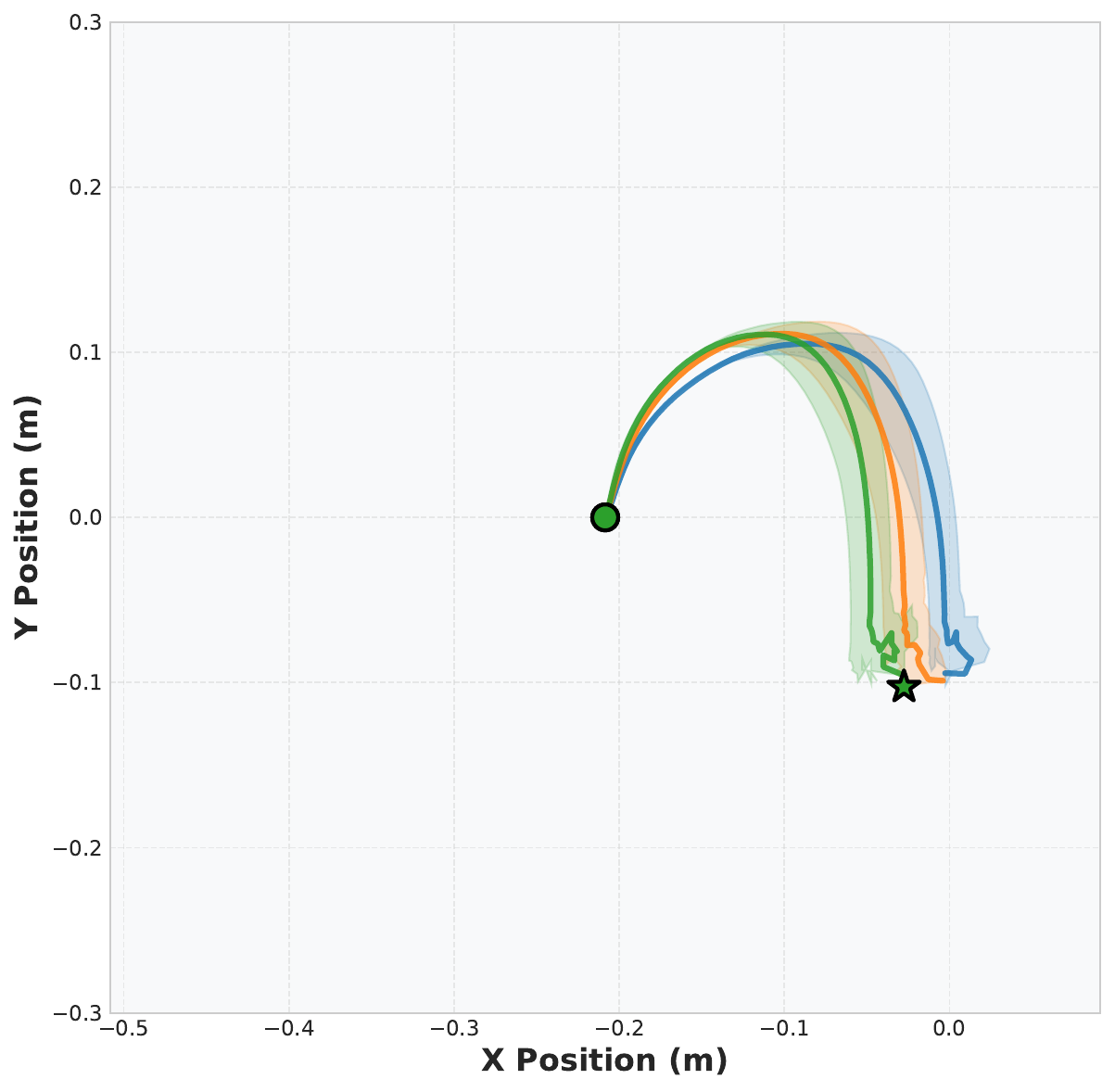}
        \caption{Sparse-only DLR (Ours)}
    \end{subfigure}
    \caption{\textbf{Top-view trajectory means.} MI-shaped training yields visually diverse but unsuccessful behaviors, while our sparse-only DLR attains success with non-trivially different trajectories across patterns.}
    \label{fig:xy_topview_mi_vs_dlr}
\end{figure}

We experimentally evaluate the full objective in Eq.~\eqref{eq:full-objective} by training MLP-Gaussian policies with an MI-shaped intrinsic reward and compare against our sparse-only DLR. We collect three skills (task IDs $\{0,1,2\}$), each with about 50 \emph{successful} episodes (\(\sim\)150 total). Episodes are padded to the maximum length with \verb|NaN|, XY is extracted from the first two state dimensions, and per-skill mean top-view trajectories are computed via \verb|nanmean| (optionally with \verb|nanstd| ribbons). The top-view visualizations (Figure~\ref{fig:xy_topview_mi_vs_dlr}) show that MI-shaped training yields trajectories that appear diverse in XY yet fail to reliably complete the drawer-closing task, whereas sparse-only DLR attains success with non-trivially different trajectories across patterns. This aligns with our analysis: on-policy MI surrogates reward discriminability on visited states and assign near-zero value to unseen states, penalizing exploration; our decoupled approach anchors diversity to a successful-state manifold and uses only sparse reward online, avoiding this failure mode.

\begin{figure}[h]
    \centering
    \includegraphics[width=\linewidth]{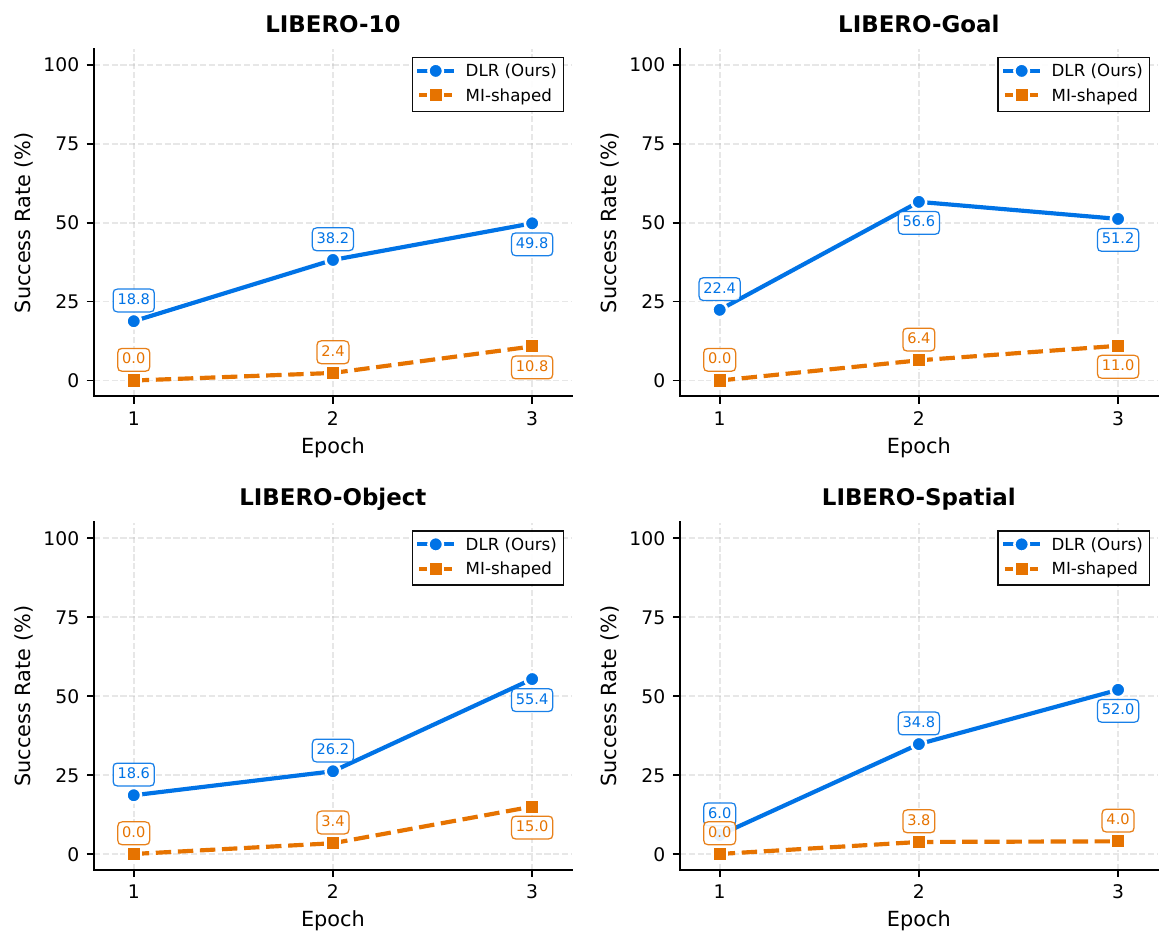}
    \caption{\textbf{Downstream adaptation vs.\ epochs.}
    Success rates on LIBERO-Long (denoted LIBERO-10), LIBERO-Goal, LIBERO-Object, and LIBERO-Spatial
    as a function of fine-tuning epochs (1--3).
    Models pretrained on DLR-generated data consistently outperform those pretrained with the MI-shaped
    objective at every epoch, demonstrating superior sample efficiency of our RL-based data generation.}
    \label{fig:epoch_comparison_all_tasks}
\end{figure}

Beyond the immediate RL behavior, we also compare how the two data-generation schemes affect downstream
VLA adaptation.
Figure~\ref{fig:epoch_comparison_all_tasks} reports the success rates on four LIBERO suites as a function
of the number of fine-tuning epochs (1--3) when pretraining is performed with DLR versus with the MI-shaped
objective.
Across all suites and at every epoch, DLR-pretrained models achieve consistently higher success rates,
with especially large margins in the early-epoch regime, indicating substantially better sample efficiency
for downstream learning.

\clearpage
\section*{D.7. LIBERO Task Suite Designs (Single-Column Visualization)}

The LIBERO benchmark~\cite{liu2023libero} consists of four primary task suites, each targeting a different aspect of generalization in visuomotor learning:
(1)~Libero-Long, 
(2)~LIBERO-Goal, 
(3)~LIBERO-Object, and 
(4)~LIBERO-Spatial.
Each suite contains a set of manipulation tasks sharing similar underlying structures but differing in visual appearance, goal condition, or spatial configuration.

\begin{figure}[h]
    \centering
    \includegraphics[width=0.95\textwidth]{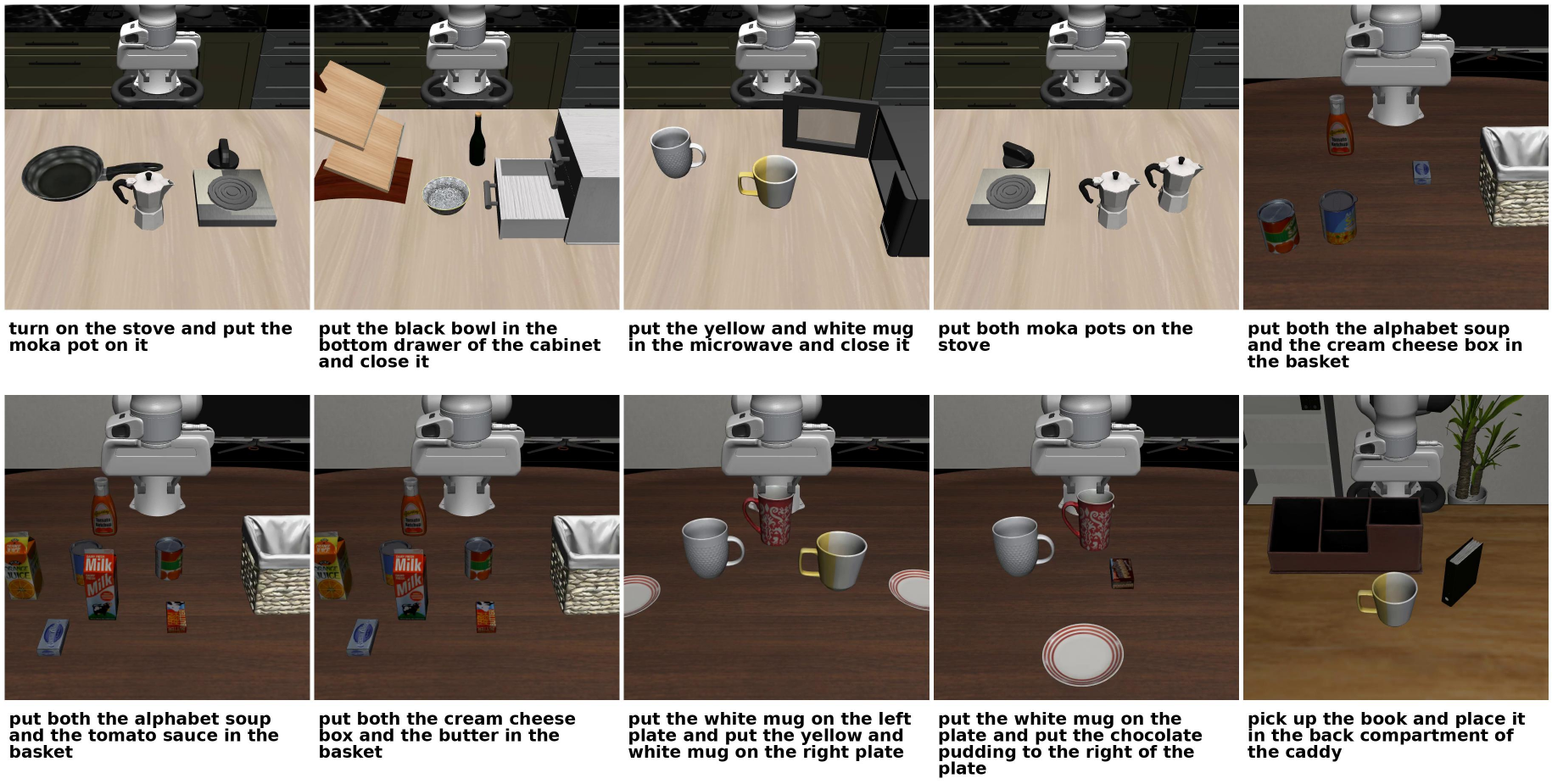}
    \caption{Visualization of the \textbf{Libero-Long} task suite. Each cell shows the initial frame and corresponding task instruction.}
    \label{fig:libero_10}
\end{figure}

\begin{figure}[h]
    \centering
    \includegraphics[width=0.95\textwidth]{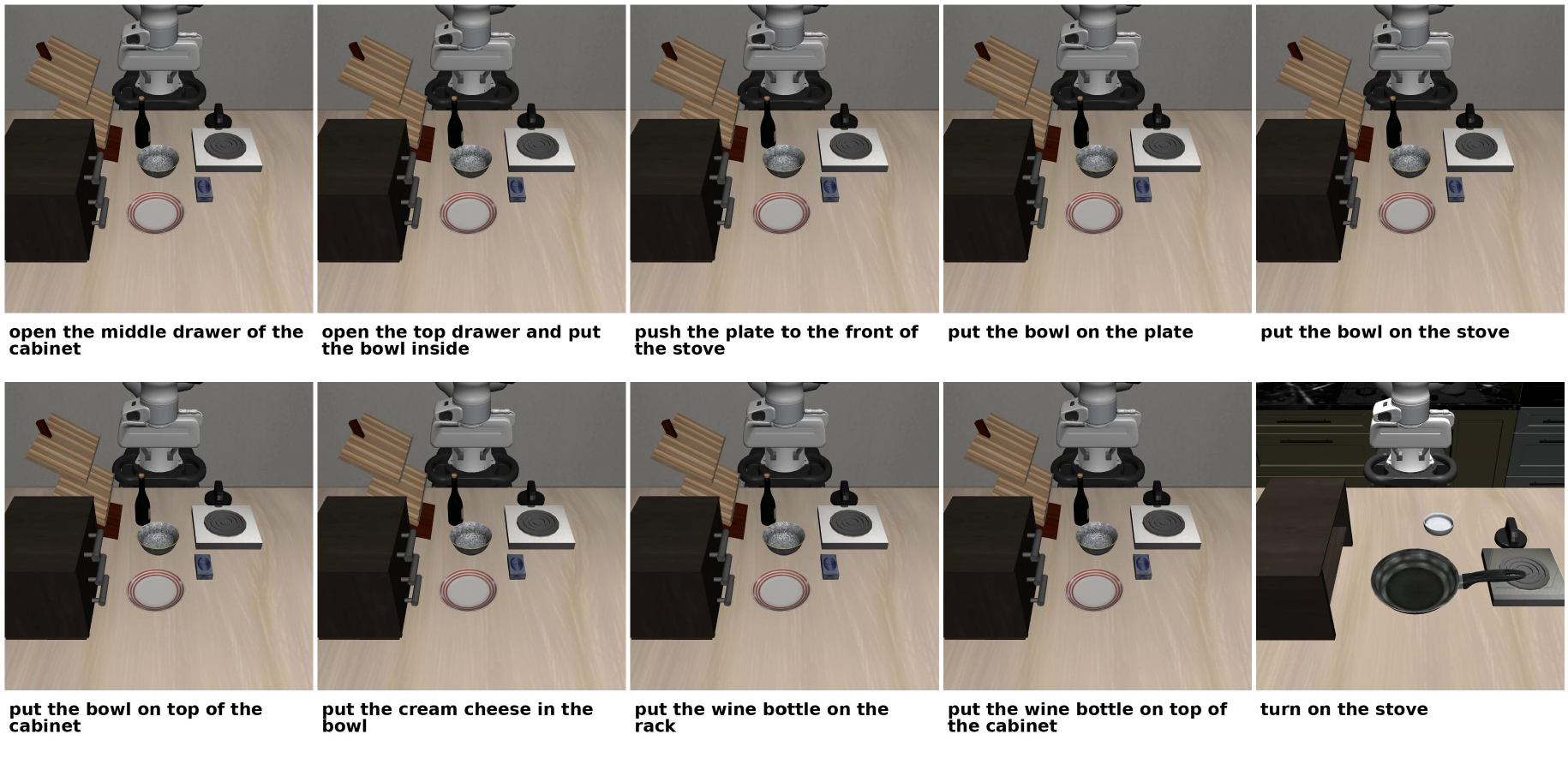}
    \caption{Visualization of the \textbf{LIBERO-Goal} suite, highlighting goal-conditioned variations.}
    \label{fig:libero_goal}
\end{figure}

\begin{figure}[h]
    \centering
    \includegraphics[width=0.95\textwidth]{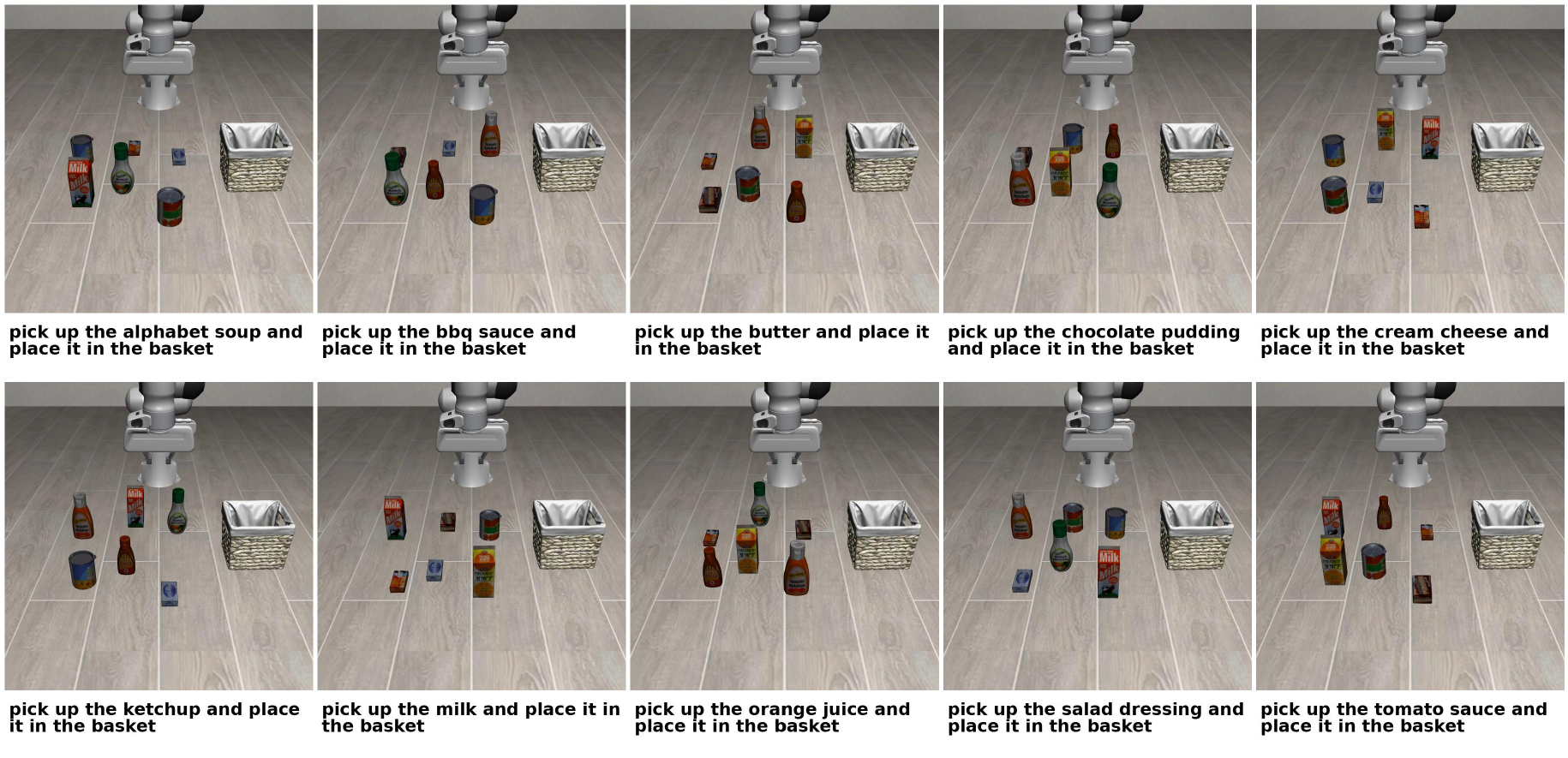}
    \caption{Visualization of the \textbf{LIBERO-Object} suite, emphasizing object diversity and category-level generalization.}
    \label{fig:libero_object}
\end{figure}

\begin{figure}[h]
    \centering
    \includegraphics[width=0.95\textwidth]{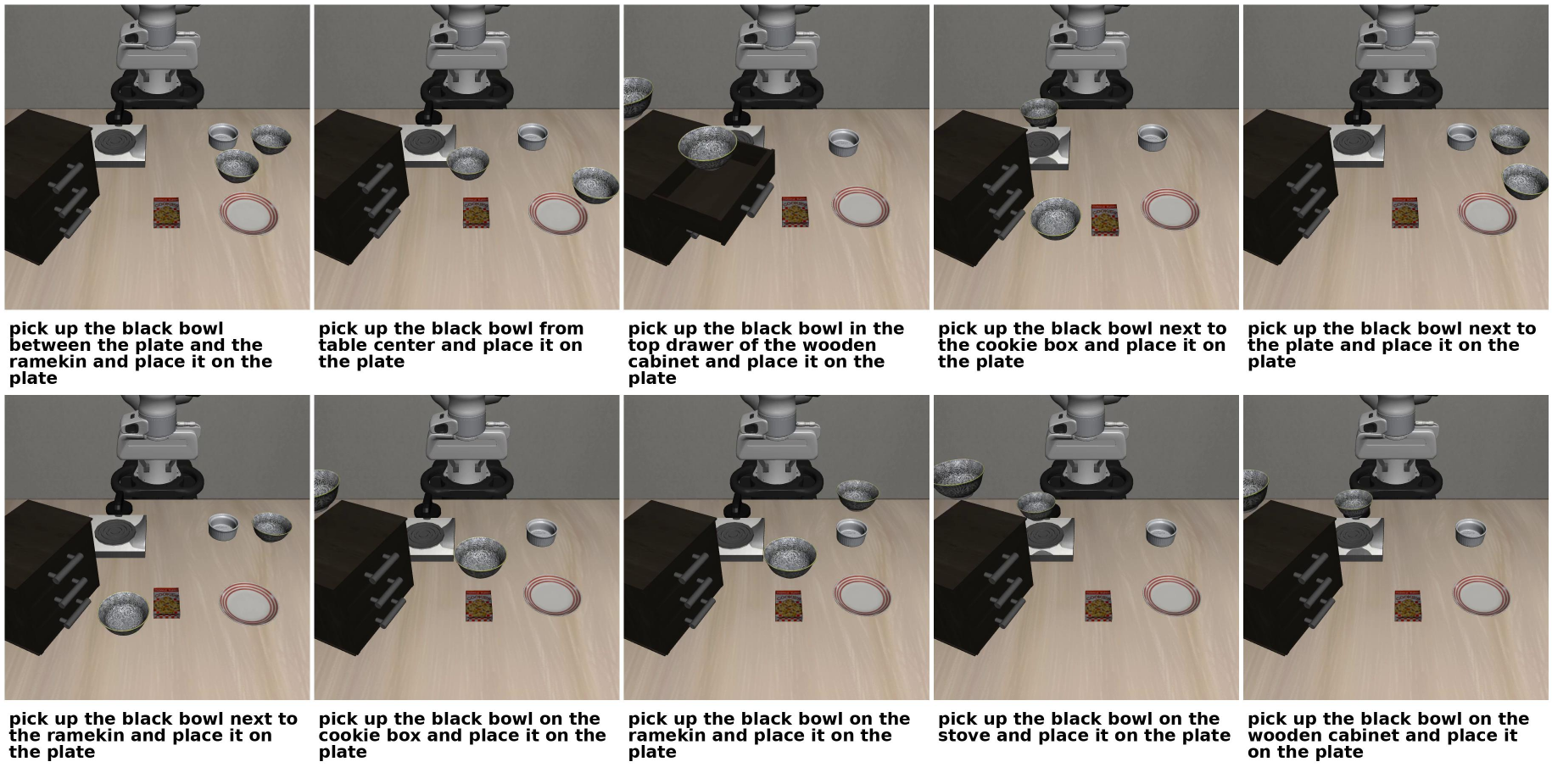}
    \caption{Visualization of the \textbf{LIBERO-Spatial} suite, designed to assess spatial reasoning and geometric consistency.}
    \label{fig:libero_spatial}
\end{figure}

Figures~\ref{fig:libero_10}–\ref{fig:libero_spatial} visualize these suites in a consistent format.
Each grid is organized as a 2$\times$5 layout, where every unit corresponds to one task.
The top image in each unit shows the first frame of the demonstration video, and the text below it displays the natural-language instruction associated with that task.
This layout emphasizes how task goals vary across different scenes while maintaining similar robotic setups.

\vspace{0.5em}

\begin{figure}[h]
    \centering
    \includegraphics[width=0.95\textwidth]{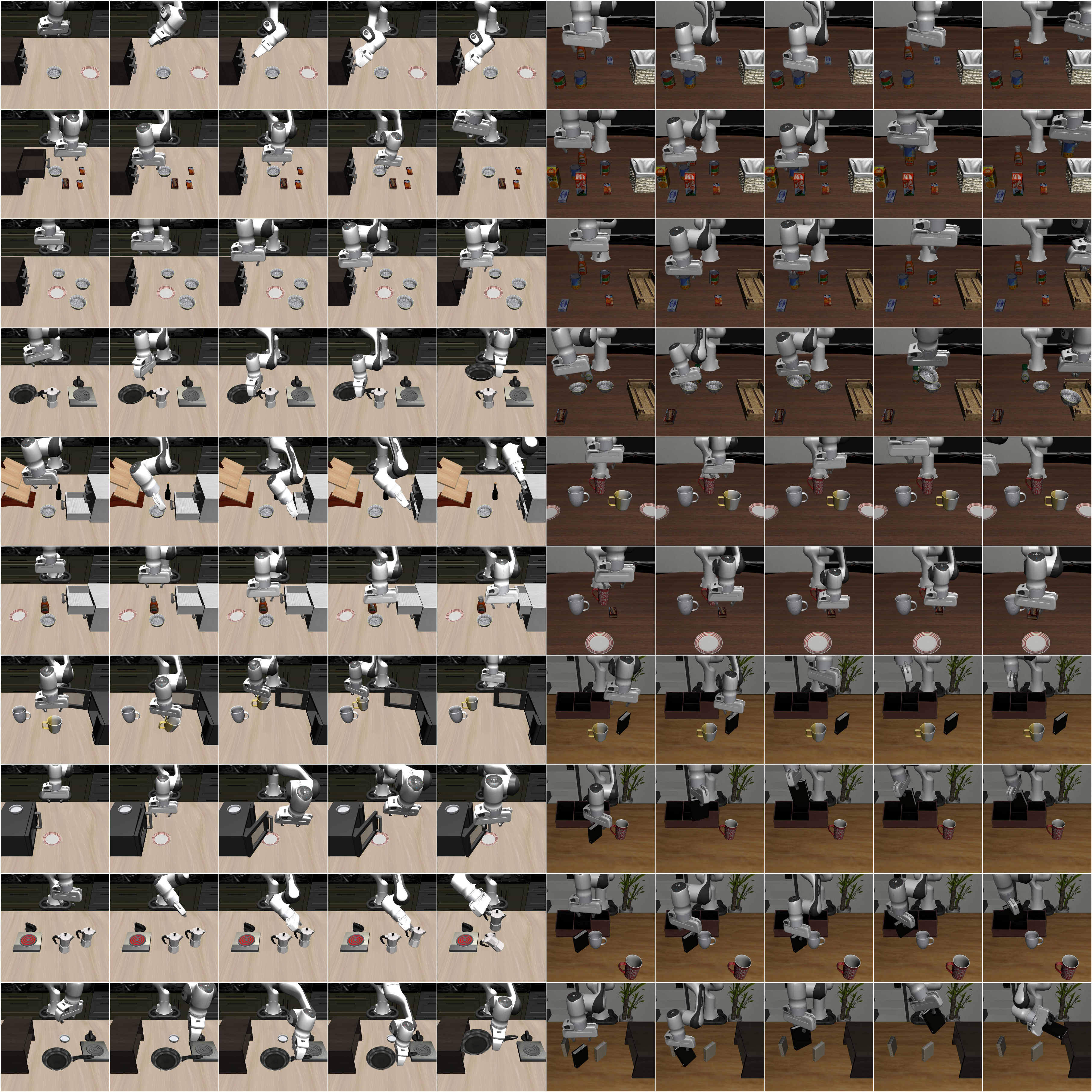}
    \caption{Comprehensive visualization of the \textbf{LIBERO-90} suite.
    Each block corresponds to one scene, displaying five evenly sampled frames.
    The grid highlights the large-scale diversity of environments and tasks in LIBERO-90.}
    \label{fig:libero_90}
\end{figure}

\vspace{1em}
\noindent
Finally, Figure~\ref{fig:libero_90} presents an overview of the comprehensive \textbf{LIBERO-90} suite.
Unlike the previous visualizations, each scene here is represented by multiple evenly sampled frames, forming a large 10$\times$10 grid (20 scenes $\times$ 5 frames per scene). 
This visualization omits text captions to focus purely on the visual and spatial diversity across the full spectrum of tasks.

\vspace{0.5em}

\end{document}